\documentclass[preprint,12pt]{elsarticle}
\usepackage{amssymb}
\usepackage{amsthm}
\usepackage{color}
\usepackage{amsmath}
\usepackage{enumitem}
\usepackage{graphicx}
\usepackage{caption}
\usepackage{booktabs}
\usepackage{subcaption}
\usepackage[T1]{fontenc}
\usepackage[utf8]{inputenc}
\usepackage{afterpage}
\usepackage[numbers]{natbib}
\usepackage{multirow}
\usepackage[hidelinks]{hyperref}
\usepackage{algorithm}
\usepackage{algpseudocode}

\theoremstyle{plain}
\newtheorem{theorem}{Theorem}

\newtheorem{definition}[theorem]{Definition}

\newtheorem{lemma}[theorem]{Lemma}

\newtheorem{remark}[theorem]{Remark}

\usepackage{amssymb}
\journal{Name of Journal}

\begin{document}

\begin{frontmatter}



\title{Generalized Naive Bayes}


\author[inst1]{Edith Alice Kovács}

\affiliation[inst1]{organization={Budapest University of Technology and Economics, Budapest, Hungary}}

\author[inst2]{Anna Ország}

\affiliation[inst2]{organization={Institute for Computer Science and Control (SZTAKI), Hungarian Research Network (HUN-REN), Budapest, Hungary}}

\author[inst1]{Dániel Pfeifer}
\author[inst2]{András Benczúr}

\begin{abstract}
In this paper we introduce the so-called Generalized Naive Bayes structure as an extension of the Naive Bayes structure. We give a new greedy algorithm that finds a good fitting Generalized Naive Bayes (GNB) probability distribution. We prove that this fits the data at least as well as the probability distribution determined by the classical Naive Bayes (NB). Then, under a not very restrictive condition, we give a second algorithm for which we can prove that it finds the optimal GNB probability distribution, i.e. best fitting structure in the sense of KL divergence. Both algorithms are constructed to maximize the information content and aim to minimize redundancy. Based on these algorithms, new methods for feature selection are introduced. We discuss the similarities and differences to other related algorithms in terms of structure, methodology, and complexity. Experimental results show, that the algorithms introduced outperform the related algorithms in many cases. 
\end{abstract}


\begin{highlights}
\item A Generalized Naive Bayes (GNB) structure is introduced;
\item It is proven that the so called GNB probability distribution associated to the GNB structure gives an approximation at least as good as the probability distribution associated to the Naive Bayes structure;
\item Two new algorithms are introduced: Algorithm GNB-A, which is an efficient greedy algorithm and Algorithm GNB-O  which finds the optimal structure given a not very restrictive condition;
\item The new algorithms' complexities are calculated and compared with the related ones;
\item Based on the algorithms a feature selection method and feature importance score are introduced. 
\item The algorithms are tested on real, medical datasets, using a "general" discretization, and compared with the related algorithms;
\end{highlights}

\begin{keyword}
Naive Bayes\sep Generalized Naive Bayes\sep conditional independence \sep  classification\sep feature selection \sep classification on health data 
%% \PACS 0000 \sep 1111
\MSC 62C12 \sep 62C10 \sep 62-07
\end{keyword}

\end{frontmatter}


\section{Introduction}
Naive Bayes is one of the most popular machine learning algorithms due to its simplicity, efficiency and easy interpretation which is appealing for experts in various domains. Therefore, Naive Bayes (NB) is considered to be one of the top 10 data mining algorithms \cite{zhang2021attribute}.

In order to show the ubiquitous interest in this algorithm we give a glimpse of some recent applications of it.
A recent work \cite{rabie2022expecting} introduces the COVID-19 Prudential Expectation Strategy (CPES) as a new strategy for predicting the behavior of a person’s body if he has been infected with COVID-19. For the classification task, they proposed the Statistical Naive Bayes algorithm. It turned out that this methodology outperforms other competing methods.  
Another paper \cite{shaban2021accurate} uses Distance Biased Naive Bayes (DBNB) classification strategy for accurate diagnosis of Covid-19 patients which combines the Weighted Naïve Bayes Module with the so-called Distance Reinforcement Module. Their experimental results showed that the introduced DBNB algorithm outperforms all competitors in maximum diagnosis accuracy and also minimum error. It is worth interesting to remark that both of these COVID-related papers started with a feature selection phase.

The paper \cite{kim2009associative} is concerned with using Naive Bayes in text mining, for automated literature research to select appropriate
gene ontology in the MEDLINE database.

The paper \cite{ye2022naive} is related to predicting lung adenocarcinoma (LUAD) from a selected gene expression. Six key genes were identified based on the tumor micro-environment, which is used as potential prognostic biomarkers and therapeutic targets of LUAD.
Using these six genes, a Naive Bayes model was constructed which predicts LUAD accurately. The predictions realized based on the Naive Bayes model were verified by the receiver operating characteristic (ROC) curve where the area under the curve (AUC) reached 92.03\%, which shows that the model used could accurately discriminate the tumor samples from the normal ones.
 
\vspace{3mm}
 
 NB algorithm is based on the assumption that the explanatory variables are conditionally independent given the classifier. This is not generally true for real-life problems.
Many papers are concerned with the improvement of this remarkable algorithm. These improvements follow mainly two directions. Some researches show that selecting some attributes beforehand might result in better classification accuracy and better generalization. Another improvement might be achieved by relaxing the conditional independence assumption.

The aim of this paper is to give a close-optimal relaxation of the conditional independence assumption at a given complexity level.

The remainder of this paper is organized as follows. In Section~\ref{sec:rel_work}, the related work is presented. Section~\ref{sec:basic_concepts} introduces the concepts that are used in our approach. In Section~\ref{sec:gnb}, the Generalized Bayes structure is introduced and discussed, how it is related to other structures which were introduced earlier. In this section, we also introduce a greedy algorithm for constructing a Generalised Naive Bayes classifier. We prove here that the introduced model gives an approximation at least as good as the original NB structure. We give a second algorithm for constructing a generalized Naive Bayes structure and prove that this gives the optimal structure in terms of fitting to the real data. In both cases, the complexity of the algorithms is calculated and compared with the related algorithms.  In Section 5, the classification process and the new feature selections are presented. In section 6 we apply the algorithm to real datasets and compare the algorithm with related algorithms like NB and TAN by using various measures like accuracy, precision, recall, F1 score, AUC. We close the paper with conclusions.

%Miért jó, alkalmazási lehetőségei- kiindulva abból, hogy a Naive Bayes mikor és mire jó. Ezt a végére!!!!!!!
\section{Related work}\label{sec:rel_work}
%Naive Bayes: ismert diskrét és folytonos formái
%TAN és ATAN
%Cseresznyefás cikkek, amire itt építünk

Naive Bayes delivers fast and effective classification with a clear theoretical foundation although its drawback is the conditional independence assumption. Many empirical comparisons between naive Bayes and decision trees such as C4.5 \cite{quinlan2014c4} show that NB predicts equally as well as C4.5. Papers like Domingos and Pazzani \cite{domingos1996beyond} and Zhang \cite{zhang2004optimality} try to explain why are Naive Bayes so effective. Domingos and Pazzani explain the good performance of NB by the 0-1 loss function, which does not penalize inaccurate probability estimations. Harry Zhang shows in \cite{zhang2004optimality} that the distribution of dependencies among all attributes over classes affects the classification of NB, not the dependencies themselves. The dependencies between attributes may exist but they cancel out. This idea is investigated under Gauss distribution. In \cite{zhang2004optimality} a sufficient condition is given under which Naive Bayes is optimal even though the conditional independence is violated.
One of the methods used to improve the performance of NB is based on attribute selection. This can be achieved by selecting subsets of explanatory variables or by weighting the features by their importance based on decision trees \cite{hall2006decision}. It is shown that using feature weights in NB improves the quality of the model compared with the original NB. In \cite{lee2011calculating} the feature weighting method was inspired by Kullback Leibler divergence.

A recent paper \cite{chen2020novel} introduces a novel selective Naive Bayes algorithm, which proposes an efficient method for selecting the attributes for calculating the class- probability. The models were built in such a way that each one is an extension of another one. The most predictive one is selected by the measures of leave-one-out cross-validation. The algorithm uses a pre-ranking of the attributes based on their mutual information with the classification variable. Empirical results show that the introduced method is superior to NB and it is comparable with Tan \cite{friedman1997bayesian} as accuracy and running time. 

Zhang and Sheng \cite{zhang2004learning} proposed a gain ratio-based feature selection algorithm. The features with a higher gain ratio are preferred.
Another way to improve the accuracy of NB is by weakening the conditional assumption. The term “semi-Naive Bayes” was introduced by Kononenko \cite{kononenko1991semi}. In this paper, conditional probabilities of joint features were computed based on the training data. The method was illustrated on medical datasets with slight improvements over the classical NB.

An important step that extends the structure of naive Bayes is the introduction of the so-called Tree Augmented Naive Bayes (Tan) \cite{friedman1997bayesian}. The algorithm assumes that the relationship among the explanatory variables is only tree-structure-like, in which the class variable directly points to all feature nodes and each feature has only one parent node from another feature. Tan algorithm showed significant improvement in accuracy compared to NB. An improvement for the TAN algorithm is presented in \cite{jiang2012improving}. This result is given by averaging multiple TAN structures, which turn out to be even more accurate than the classification using only one TAN structure.

Very recent papers like paper \cite{zhang2021attribute} and \cite{wang2022alleviating} introduced new algorithms for classifications. The algorithm in paper \cite{zhang2021attribute} is based on attribute weighting and instance weighting. They define a new weighting methodology that uses the correlation of the attributes with the class variables and also with the other attributes. The weight of an attribute is defined as the difference between relevance and average redundancy. In paper \cite{wang2022alleviating} the author argues that one-dependence estimator (AODE) performs extremely well against more sophisticated models, therefore they improve the model by a special weighting based on mutual information.

A very important paper that uses decomposable models to expand the classical NB was published in Ghofrani et al. in 2018 \cite{ghofrani2018new}. This work is also interesting because, in their comparisons with other models, they use cherry tree probability distribution defined on the explanatory attribute set. They concluded that cherry trees were not so efficient because of the overfitting. The basic idea of their paper related to cherry trees is to replace the tree between the explanatory variables given by the TAN algorithm, with a cherry tree. They introduce also their model, which aims to capture interdependencies between the explanatory attributes. Their experimental performance shows that their algorithm had a better classification performance than, NB and Cherry tree-based ones. In Section~\ref{sec:gnb} we discuss the similarities and the main differences between the algorithms we introduce in this paper and those introduced earlier in \cite{ghofrani2018new}, TAN in \cite{friedman1997bayesian}, ATAN  \cite{jiang2012improving} and LDMLCS \cite{ghofrani2018new}.

The new algorithms, we introduce in this paper, are structure-extending algorithms that alleviate the conditional independence assumption of the NB. Our algorithms' output can be used as the base for a new feature selection method. We prove that our so-called Generalized Naive Bayes structure is at least as good as the NB structure in terms of fitting to the real data. We give two algorithms. The first one is a good fitting greedy algorithm, the second one gives as output the optimal structure, under a not very restrictive condition, which is rigorously proven.

\section{Basic concepts}\label{sec:basic_concepts}
Naive Bayes-related studies approach the classification problem from directed probabilistic networks, called Bayes Networks. In this paper, we will improve the naive Bayes structure by using the undirected probabilistic graphical models which are called also Markov networks, and we are concerned with an important subclass of them called cherry junction trees.

Probabilistic graphical models bring together graph theory, probability theory, and also in some cases information theory, in order to provide a probabilistic flexible framework for modeling random variables with complex interactions. However, structure-learning for graphical models remains an open challenge, since many times one must cope with a combinatorial search over the space of all possible structures. In this paper we define a special family as a search space.

At the heart of probabilistic graphical models stands the conditional independence concept (CI). In 1950 Loeve \cite{loeve1955probability} defined the concept of CI in terms of $\sigma -$algebras. The properties of CI were formulated in different contexts as follows. Phi Dawid formulated the formal properties of CI from a statistical point of view \cite{dawid1979conditional}. Later Spohn \cite{spohn1980stochastic} formulated the properties of CI in a philosophical logic context and also related CI to the concept of causality.

The following subsection contains the concepts and preliminary results, which will be used in this paper.

\subsection{Probabilistic graphical models used in the GNB structures}
First, we define the cherry tree graph structure. This concept was originally named $t$-cherry in \cite{kovacs2010approximation},  but since it does not lead to any confusion we call it simply a cherry tree graph.

We define first the cherry tree graph of order $k$ by construction.

\begin{figure}
    \centering
    \includegraphics[width=0.9\textwidth]{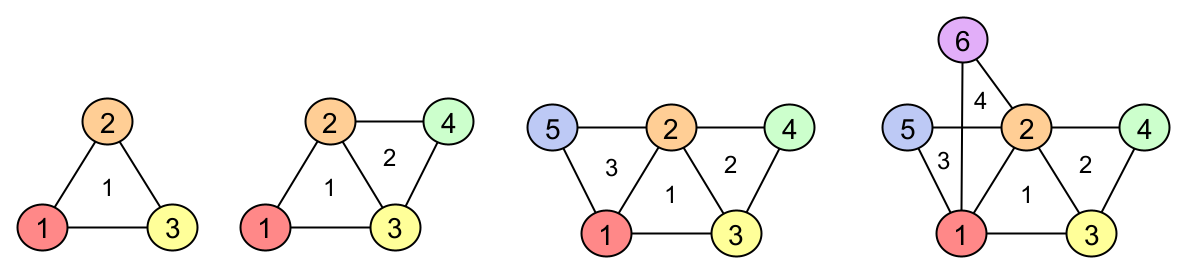}
    \caption{An example construction of a third-order cherry tree.}
    \label{fig:cherry_example}
\end{figure}

\begin{definition}
\label{$k$-th order cherry tree}
We call \textbf{cherry tree graph structure of order $k$}, a graph structure obtained by the following two steps:
\begin{enumerate}
    \item The smallest cherry tree consists of $k$ interconnected vertices.  
    \item A cherry tree on $m+1$ vertices is obtained from a cherry tree on $m$ ($m>k$) vertices by connecting a new vertex to $k-1$ already connected vertices.
\end{enumerate} 
\end{definition}
Figure \ref{fig:cherry_example} shows how a third-order cherry tree on 6 vertices is built.

\begin{remark}
\label{remark2}
A cherry tree is a triangulated (chordal) graph. 
\end{remark}
The following definition is a consequence of Remark \ref{remark2}.

\begin{figure}[h]
    \centering
    \includegraphics[width=0.9\textwidth]{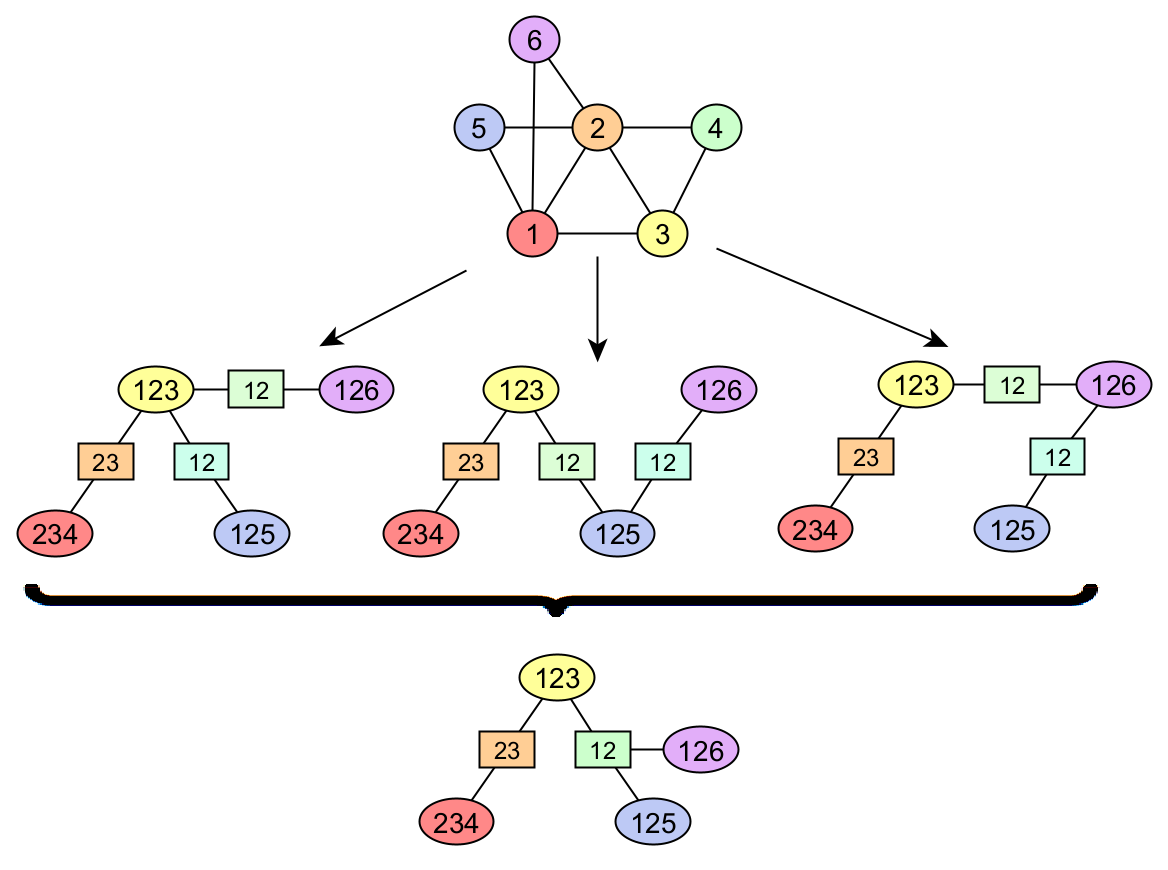}
    \caption{Cherry tree structure represented as a chordal graph (first row), represented as junction tree (second row) and represented in a compact junction tree form (third row)}
    \label{fig:cherry_junction}
\end{figure}

\begin{definition}
\label{cherry-junction tree}
We call \textbf{cherry-junction tree of order $k$} the structure assigned to a cherry tree of order $k$ in the following way:

\begin{enumerate}[label=(\roman*)]
\item To the set of vertices of each $k$-th order maximum clique a so-called \emph{cluster} is assigned ;
\item A so-called edge connects two $k$ -element-clusters if the following two conditions are fulfilled: 

- the clusters share $k-1$ elements

- if a set of $k-1$ elements is contained by $m$ clusters, these clusters will be connected tree-like by $m-1$ edges, see second row of Figure \ref{fig:cherry_junction}

\item The $k-1$-element set given by the intersection of two connected clusters is called a \emph{separator}.
\end{enumerate}
\end{definition}

For an illustration of a 3-rd order cherry-junction tree, corresponding to a 3-rd order cherry tree graph, see Figure \ref{fig:cherry_junction}.

\begin{remark}
    The junction tree associated with a cherry tree graph structure is not unique.
\end{remark}

\begin{remark}
\label{running_int_prob}
    Any junction tree satisfies the so-called ‘‘running intersection property'', i.e.\ a vertex is contained in two different clusters it is contained in any cluster on the path between the two clusters.
\end{remark}
For a good overview of the connection between these concepts, see \cite{pfeifer2024vine}.

\vspace{2mm}

Let $\mathbf{X}=\left( X_{1},\ldots ,X_{d}\right)^{T}$ \ a random vector
and  $V$ $=\{1,...,d\}$ the set of indices, and let us consider a junction tree over $V$ given by the set of clusters $\mathcal{C}$ and the set of separators $\mathcal{S}$.

\begin{definition} 
\label{def:cherry_tree_pd}
\textbf{The cherry tree probability distribution} corresponding to a cherry-junction tree $(V,\mathcal{C},\mathcal{S})$ is given by the following formula:
\begin{equation}
\label{eq:Cherry_tree}
P_{cherry}\left( X\right) =\frac{\prod\limits_{C\in \mathcal{C}}P\left( \mathbf{X}_{C}\right) }{\prod\limits_{S\in \mathcal{S}}P\left( \mathbf{X}_{S}\right)
^{v_{S}-1}}
\end{equation}
where $v_{S}$ is the number of clusters that are connected through the
separator $S$.  
\end{definition}
\begin{remark}
It is easy to see that regardless of which graphical representation of a junction tree we use (see for example Figure \ref{fig:cherry_junction}) the cherry tree probability distribution formula is the same:
\[
P_{cherry}\left( X\right) =\frac{P( \mathbf{X}_{123}) P( \mathbf{X}_{234}) P( \mathbf{X}_{125})P( \mathbf{X}_{126})}{(P\left( \mathbf{X}_{12}\right))
^{2}P\left( \mathbf{X}_{23}\right)}.
\]
\end{remark} 
\vspace{3mm}
In this paper, the generalization of the NB structure is derived from probabilistic graphical models.
Therefore after we define the probability distribution assigned to an NB structure we will express it in terms of junction trees.
\begin{definition}
\label{def:NB_pd}
The \textbf{Naive Bayes probability distribution}, when the attributes $X_i$ are supposed to be conditionally independent given the class variable $Y$, has the following expression:
\[ 
   P_{NB}(\mathbf{X},Y)= P(Y){\prod%
\limits_{i=1}^{d}P(X_{i}\mathbf{|}Y)}
\]
\end{definition}

\begin{remark}
    It is easy to see that Definition \ref{def:NB_pd} is equivalent to the following formula:
    \begin{equation}
    \label{eq:NB_equivalent_formula}
        P_{NB}(\mathbf{X},Y)=\frac{\prod
        \limits_{i=1}^{d}P(X_{i}\mathbf{,}Y)}{P(Y)^{d-1}}
    \end{equation}
\end{remark}
In terms related to junction trees, the clusters consist of two elements ${X_i,Y}$, $i=1,\dots,d$ and the separators consist of only one element, namely ${Y}$.

\vspace{2mm}

The Naive Bayes classifier assigns to a new realization $\textbf{x}$ the most probable value $y\left( \mathbf{x}\right)$ of the class variable defined in the following manner.
\[
y\left( \mathbf{x}\right) =\underset{y_{i}}{arg\max }\left( p_{NB}(y_{i},%
\mathbf{x}\right)) 
\]
where $y_i,\ i=1,\dots,s$ denote the classes of the class variable $Y$. 
It is clear that this classifier uses conditional independencies between the explanatory variables given the class variable. These conditional independencies are usually in real-life problems not satisfied, it seems to be conservative. This fact motivates our research too.

\subsection{Information theoretical concepts}
In this subsection, we summarize the main concepts of information theory (see for example the fundamental book \cite{cover2012elements}).

We define conditional entropy with an equivalent formulation which we will use in this paper.
\begin{definition}
The \textbf{conditional entropy} quantifies the amount of information needed to describe the outcome of a random variable $Y$  given that the value of another random variable $X$ is given by
\[
H(Y|X)=H(X,Y)-H(X).
\]
\end{definition}

\begin{definition}
The \textbf{information content} of a random vector $\mathbf{X}=(X_{i_{1}},...,X_{i_{k}})^{T}$ is given by
\[
I(X_{i_{1}},...,X_{i_{k}})=\sum_{\mathbf{x}}P\left( x_{i_{1}},\dots,x_{i_{k}}\right) \log \frac{P\left( x_{i_{1}}\mathbf,\dots,x_{i_{k}}\right) }{P(x_{i_{1}})\cdot \ldots \cdot P(x_{i_{k}})}.
\]
\end{definition}
 The information content can be seen as a generalization of the mutual information. There are also other generalizations for the mutual information.
If $P(\mathbf{X})$ is approximated by a probability distribution $P_{app}(\mathbf{X})$, a quantification of the goodness of the approximation is given by Kullback and Leibler in 1951 \cite{kullback1951information}.

\begin{definition}
The following expression is called \textbf{Kullback-Leibler divergence} between a given joint probability distribution $P(\mathbf{X})$ and its approximation $P_{app}(\mathbf{X})$:
\[
\mathrm{KL}(P_{app}(\mathbf{X}),P(\mathbf{X}))=\sum_{\mathbf{x}}P\left( \mathbf{x}\right) \log \frac{P\left( \mathbf{x}\right) }{P_{app}(\mathbf{x})}.
\]
\end{definition}

\begin{remark}
Kullback-Leibler divergence (KL divergence) is a positive number unless $P_{app}\equiv P$, when $\mathrm{KL}=0$. The smaller the value of the KL-divergence is the better the approximation is.
\end{remark}

A fundamental theorem we use in this paper expresses the KL divergence between a probability distribution and a cherry-tree probability distribution approximation (\ref{eq:Cherry_tree}).

\begin{theorem}\cite{szantai2012hypergraphs}
The Kullback-Leibler divergence between the cherry tree approximation (\ref{eq:Cherry_tree}) and the real distribution $P(\mathbf{X})$ is:
\begin{multline}
\label{eq:kl_div}
\mathrm{KL}\left(P_{\it{Cherry}}(\mathbf{X}),P(\mathbf{X})\right)=\\
-H(\mathbf{X})-\left[\sum_{\mathbf{X}_C\in\mathcal{C}}I(\mathbf{X}_C) -\sum_{\mathbf{X}_S\in\mathcal{S}}(\nu_S-1)(I(\mathbf{X}_S))\right]+\sum_{i=1}^nH(X_i)
\end{multline}
\end{theorem}

These were the basic information theoretical concepts that we will use in the next part.

\section{Generalized Naive Bayes}\label{sec:gnb}
This part contains the new results of the paper.
First, let us introduce the Generalized Naive Bayes structure, which is an NB structure augmented with certain edges.

As we saw earlier the naive Bayes structure can be assigned to a junction tree with two elements in each cluster see \ref{eq:NB_equivalent_formula}.

The new concept introduced in this paper is basically a generalization of the above idea, i.e., we will use special junction trees, with exactly three elements in a cluster (two besides $Y$) and two elements in the separator. This way we will use a cherry tree structure for the generalization.

\subsection{Generalized Naive Bayes graph structure}

\bigskip We introduce the Generalized Naive Bayes graph structure and then define the probabilistic graphical model associated with it.

\begin{definition}\label{def:GB_gstructure}
A  \textbf{Generalized Bayes graph-structure} on $V\cup\left\{0\right\}=\\$ $ =\left\{ 0,1,\ldots ,d\right\}$ (vertex 0 is assigned to $Y$), is constructed as follows.
\begin{itemize}
\item Step 1. The smallest GNB structure on three vertices is given by the interconnected triplet $\left(
0,i_{1},i_{2}\right) $, where $i_{1},i_{2},\in V$. 
\item Step k. Let us suppose that the vertices $i_{1},i_{2},\ldots
,i_{k}\in V$ are already in the GNB structure. We add a new vertex $i_{k+1}\in V$ to the existing GNB structure by connected it to vertex $0$ and an already connected
vertex $i_{m}\in \left\{ i_{1},i_{2},\ldots ,i_{k}\right\} $. The vertex $i_m$ is
called the  \emph{mother of vertex} $i_{k+1}$, and it is denoted by $\mu (i_{k+1})$. Vertex $0$ is called the  \emph{father vertex} of the new $(i_{k+1})$ vertex.
\end{itemize}
\end{definition}

\begin{remark}
It is easy to see that a GNB is a special type of cherry-junction tree structure (See Definition \ref{cherry-junction tree})
\end{remark}

\begin{remark}
    Each vertex $i_k, k>1$, has one mother vertex $\mu (i_k)$ and the same father vertex $0$. The first connected vertex $i_1$ has only the father vertex $0$.
\end{remark}

\begin{figure}
    \centering
    \includegraphics[width=0.5\textwidth]{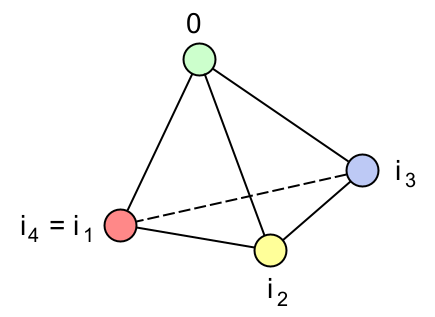}
    \caption{The maximum clique is of size four, which leads to a contradiction}
    \label{fig:maxclique_size4}
\end{figure}

\begin{figure}
    \centering
    \includegraphics[width=0.6\textwidth]{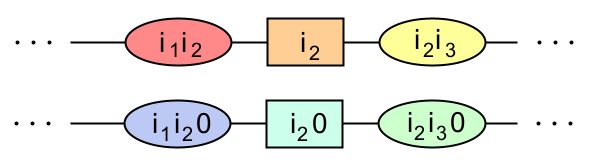}
    \caption{Adding the vertex $0$ to each cluster}
    \label{fig:implication_proof_junction}
\end{figure}

An important remark, that connects this work to earlier research is the following.
\begin{theorem}
\label{tree between X}
A graph structure defined on the vertex set $\left\{ 0,1,\ldots ,d\right\}$ has a GNB structure (Definition \ref{def:GB_gstructure}) if and only if it has the property that its restriction to the vertices  $\left\{ 1,\ldots ,d\right\} $ form a tree.
\end{theorem}
\begin{proof}
First we prove that a GNB on $\left\{ 0,1,\ldots ,d\right\}$ implies the tree structure on $\left\{ 1,\ldots ,d\right\}$.

    We do the proof by reductio ad absurdum. 
    Let us suppose that the vertices in $\left\{ 1,\ldots ,d\right\}$ in a GNB graph structure are not connected tree-wise. This means that there exists a path $\left\{ i_{1},i_{2},\ldots ,i_{r}=i_{1}\right\}$, $r>3$, where $i_{1},i_{2},\ldots ,i_{r}\in\ V$.
    
    We prove this by considering the following two cases:
   \begin{itemize}
       \item if $r=4$ then $0, i_{1}, i_{2}, i_{3}, i_{4}$, are interconnected, which implies that the maximum clique is of size four (see Figure \ref{fig:maxclique_size4}), which is in contradiction, with Definition \ref{def:GB_gstructure}.
       \item if $r>4$ then we have at least three connected triplets (which share two common vertices): $\left( 0,i_{1},i_{2}\right)$,  $\left( 0,\mu(i_3),i_{3}\right)$, $\left( 0,\mu(i_4),i_{4}\right)$, where $\mu(i_3)$ is the mother vertex of $i_3$, and $\mu(i_4)$ is the mother vertex of $i_4$, and $\mu (i_{3})\in \left\{ i_{1},i_{2}\right\} $, $\mu (i_{4})\in \left\{
i_{1},i_{2},i_{3}\right\} $
       Since $i_1=i_r$ it belongs to the first and last triplet, and obviously it does not belong to the intermediate triplets. Therefore the running intersection property is not fulfilled.
 \end{itemize}
       Now we will prove the other implication.
       Let us consider a tree defined on the vertices  $\left\{ 1,\ldots ,d\right\}$. This tree can be expressed in terms of junction tree, with two vertices in each cluster, and one vertex in each separator, see the first row in Figure \ref{fig:implication_proof_junction}. Then we add to each cluster the vertex $0$, because this is connected by definition with all vertices  $\left\{ 1,\ldots ,d\right\}$. It is easy to see that the new junction tree will have three vertices in a cluster and two in each separator (the old vertex and the vertex $0$, which is now in all the clusters (see Figure \ref{fig:implication_proof_junction}, the second row).
 
\end{proof}

\begin{definition}
The permutation  $i_{1},i_{2},\ldots,i_{d}$ of the indices $\left\{ 1,\ldots,d\right\}$, given by the steps of Definition \ref{def:GB_gstructure} is called  \textbf{$c$-ordering}, where the first two indices in the smallest cherry tree are arranged increasingly.
\end{definition}

Without loss of generality, we denote a $c$-ordering $i_{1},i_{2},\ldots,i_{d}$ by $1,\ldots,d$.

\begin{definition}
\label{GNB_pd}
The  \emph{GNB probability distribution} associated to the GNB graph structure and a $c$-ordering $1,\ldots,d$ is the following:
\[
P_{GNB}=\frac{\prod\limits_{i=2}^{d}P\left( Y,X_{\mu
(i)},X_{i}\right) }{\prod\limits_{i=3}^{d-1}P\left( Y,X_{_{\mu
(i)}}\right) }
\]
\end{definition}

Now we want to prove that a GNB probability distribution given by Definition \ref{GNB_pd} gives a better approximation than the NB probability distribution, in terms of Kullback- Leibler divergence.
For this, we need the Expressions of the Kullback- Leibler divergence of the NB, and GNB approximations.
\begin{theorem}
\label{KL_NB_th}
If the joint probability distribution $P(\mathbf{x,}y)$ is approximated by a Naive Bayes distribution (\ref{eq:NB_equivalent_formula}) the Kullback-Leibler divergence between the true distribution and the approximation is given by
the formula:
\begin{equation}\label{Kl_NB_amit biz kell}
    KL(P(\mathbf{x,}y),P_{NB}(\mathbf{x},y))=\sum%
\limits_{i=1}^{d}H(X_{i})+H(Y)-H(\mathbf{X,}Y)-\sum%
\limits_{i=1}^{d}I(X_{i},Y)
\end{equation}
\end{theorem}

\begin{proof}
\begin{multline}
KL(P(\mathbf{x,}y),P_{NB}(\mathbf{x},y)) = \sum\limits_{\left( \mathbf{x},y\right) }P(\mathbf{x,}y)\log \frac{P(\mathbf{x,}y)}{P_{NB}(\mathbf{x},y)}\\
=\sum\limits_{\left( \mathbf{x},y\right) }P(\mathbf{x,}y)\log P(\mathbf{x,}y) - \sum\limits_{\left( \mathbf{x},y\right) }P(\mathbf{x,}y)\log P_{NB}(%
\mathbf{x},y). \label{KL_NB}
\end{multline}

The first term in (\ref{KL_NB}) is $-H(\mathbf{X},Y)$. We calculate now the second term by exploiting the conditional independence relation i.e.:

\begin{eqnarray*}
\sum\limits_{\left( \mathbf{x},y\right) }P(\mathbf{x,}y)\log P_{NB}(\mathbf{x},y) &=&\sum\limits_{\left( \mathbf{x},y\right)}P(\mathbf{x,}y)\log \frac{\prod\limits_{i=1}^{d}P(x_{i}\mathbf{,}y)}{P(y)^{d-1}} \\
&=&\sum\limits_{\left( \mathbf{x},y\right) }P(\mathbf{x,}y)\log\prod\limits_{i=1}^{d}P(x_{i}\mathbf{,}y)-\sum\limits_{\left( \mathbf{x},y\right) }P(\mathbf{x,}y)\log P(y)^{d-1} \\
&=&\sum\limits_{\left( \mathbf{x},y\right) }P(\mathbf{x,}y)\log\prod\limits_{i=1}^{d}\frac{P(x_{i}\mathbf{,}y)}{P(x_{i})P(y)} \\
& & +\sum\limits_{\mathbf{x}}P(x_{i})\log P(x_{i})+P\left( y\right) \log P(y) \\
&=&\sum\limits_{i=1}^{d}I\left( X_{i}\mathbf{,}Y\right) -\sum\limits_{i=1}^{d}H(X_{i})-H\left( Y\right)
\end{eqnarray*}

Now we substitute this result in the formula (\ref{KL_NB}), and we get the formula (\ref{Kl_NB_amit biz kell}).
\end{proof}

It is easy to see that this is in fact a consequence of the Chow-Liu tree method \cite{chow1968approximating}.

\begin{remark}
The goodness of fit depends on the $%
\sum\limits_{i=1}^{d}I\left( X_{i}\mathbf{,}Y\right) $. As larger this sum is the better the approximation is.
\end{remark}
This result highlights why the Naive Bayes works so well, and why it is worth choosing the explanatory variable based on their mutual information with the classifier.
\begin{theorem}
\label{KL_GNB_th}
If  $P(\mathbf{X})$ is approximated by a Generalized Naive Bayes
distribution $P_{GNB}(\mathbf{X})$ (see Definition \ref{GNB_pd}) then the KL divergence is given by the
formula:%
\begin{eqnarray*}
KL(P(\mathbf{X,}Y),P_{NB}(\mathbf{X},Y))
&=&\sum\limits_{i=1}^{d}H(X_{i})+H(Y)-H(\mathbf{X,}Y) \\
&&-\left( \sum\limits_{i=2}^{d}I\left( Y,X_{\mu (i)},X_{i}\right)
-\sum\limits_{i=3}^{d}I\left( Y,X_{\mu (i)}\right) \right) 
\end{eqnarray*}
\end{theorem}

\begin{proof}
This follows straightforwardly from the formula of KL divergence applied for an approximation given by a cherry tree probability distribution, given by formula (\ref{eq:kl_div}).
\end{proof}

We saw in Theorem \ref{KL_NB_th} and Theorem \ref{KL_GNB_th} that the corresponding KL divergences depend on two parts, one of them being common to both approximations. Therefore the goodness will depend on the part that differs from one model to another, from NB to GNB.

\begin{definition}
\label{def:weight_of_approx}
Let $P_{app}(\mathbf{X})$ be an approximation for $P(\mathbf{X})$. We call the  \textbf{weight of the approximation} 
\[
W(P(\mathbf{X}),P_{app}(\mathbf{X}))=\sum_{i=1}^{d}H\left( X_{i}\right)
-H\left( \mathbf{X}\right) -KL\left( P(\mathbf{X}),P_{app}(\mathbf{X}%
)\right).
\]
\end{definition}

\begin{remark}
       The weight of the $P_{NB}P(\mathbf{X},Y)$ approximation is:
\[
W(P(\mathbf{X},Y),P_{NB}P(\mathbf{X},Y))=\sum\limits_{i=1}^{d}I\left(Y,X_{i}\right).
\]
\end{remark}

\begin{remark}
The weight of the $P_{GNB}P(\mathbf{X},Y)$ approximation is:
\begin{eqnarray}
\label{Weight_GNB}
 W(P(\mathbf{X},Y),P_{GNB}P(\mathbf{X},Y))=
 \sum\limits_{i=2}^{d}I\left( Y,X_{\mu (i)},X_{i}\right)
-\sum\limits_{i=2}^{d}I\left( Y,X_{\mu (i)}\right) \ 
\end{eqnarray}
\end{remark}

Now we will show that the weight of a GNB approximation is at least as large as the weight of an NB approximation. To prove this we need first the following two theorems.
\begin{theorem}
The weight of the NB approximation can be written as:
\begin{equation}
\label{eq:weight_of_NB}
W(P,P_{NB})=\sum\limits_{i=1}^{d}H\left( X_{i}\right)
-\sum\limits_{i=1}^{d-1}H\left( X_{i}|Y\right) -H\left( Y,X_{d}\right)
\end{equation}
\end{theorem}

\begin{proof}
\begin{eqnarray*}
W(P,P_{NB}) &=&\sum\limits_{i=1}^{d}I\left( Y,X_{i}\right) \\
&=&H\left( Y\right) +H(X_{1})-H(Y,X_{1}) \\
&&+H\left( Y\right) +H(X_{2})-H(Y,X_{2}) \\
&&+H\left( Y\right) +H(X_{3})-H(Y,X_{3}) \\
&&\vdots \\
&&+H\left( Y\right) +H(X_{d-1})-H(Y,X_{d-1}) \\
&&+H\left( Y\right) +H(X_{d})-H(Y,X_{d})
\end{eqnarray*}

First we add the entropies $H(Y)$, $H(X_{1})$,$H(X_{2})$,\ldots ,$H(X_{d})$,
then we apply the formula $H(Y,X_{i})-H\left( Y\right) =H(X_{i}|Y)$ starting
on from the first row, for $i=1,\ldots d-1$. This way we get the formula (\ref{eq:weight_of_NB}).
\end{proof}

To prove a similar theorem regarding the weight of a GNB approximation, we need a result related to the GNB graph structure.

First, we define the concept of a chain of clusters.
\begin{definition}
 A \textbf{chain of clusters} is a special cherry tree in which each of the clusters is connected to at most two different clusters.   
\end{definition}
\begin{remark}
    A chain cluster has exactly two simplicial vertices (vertices connected only to the other two (in the general case, to $k-1$ ) vertices)
\end{remark}
\begin{lemma}
\bigskip A GNB graph structure can be represented as a chain of clusters.
\end{lemma}

\begin{proof}
Let us suppose that we have a GNB structure with the property that there
exists a cluster that is connected to three clusters by different
separators. We denote this cluster by $C_{1}\left( 0,i,j\right) $ and its
three neighbours by $C_{2},C_{3},C_{4}$. This implies that it is possible to connect $C_{1}$with two clusters by two different separators $S_{12}$ $%
\left( 0,i\right) $ and $S_{13}$ $\left( 0,j\right) $, but we have no other different subset in $C_{1}$ which contains the vertex  $0$ (which
corresponds to class variable $Y$). Therefore any cluster can be connected
to the other clusters by at most two different separators, which completes
the proof.
\end{proof}

Since the GNB probability distribution associated to the GNB graph can be
defined as a junction tree associated to a chain of clusters, we will assign a numbering to the random variables involved.

\begin{definition}
The chain numbering of the attributes based on the chain structure of GNB  is defined as follows. $C_{1},C_{2},\ldots ,C_{d-1}$ are the clusters of the chain, $%
C_{1}=( Y,X_{i_{1}},X_{i_{2}}) $, $C_{2}\backslash C_{1}$is
denoted by $X_{i_{3}}$, \ldots , $C_{t}\backslash C_{t-1}$ is denoted by $%
X_{i_{t+1}}$, \ldots ,$C_{d-1}\backslash C_{d-2}$ is denoted by $X_{i_{d}}$.
(Where $"A\backslash B"$ denotes the elements belonging to the set $A$ and not to the set $B$.)
\end{definition}

\begin{remark}
The chain numbering $\left\{X_{i_{1}}, \ldots , X_{i_{d}}\right\} $ is given by a permutation of the indices of the attributes and it is not unique. 
\end{remark}

For simplicity without loss of generality, we
will denote this numbering $\left\{ i_{1},\ldots ,i_{d}\right\} $ by $% 
\left\{ 1,\ldots ,d\right\} $ but whenever we use this we mention that this is a chain numbering.

\begin{theorem}
Let  $\left\{ 1,\ldots ,d\right\} $ be a chain numbering. The weight of the basic GNB approximation is given by the formula:%
\begin{equation}\label{keplet}
W_{GNBbasic}^{\left( d\right)
}=\sum\limits_{i=1}^{d}H(X_{i})+H(Y)-\sum%
\limits_{i=1}^{d-1}H(X_{i}|Y,X_{i+1})-H(Y,X_{d}).
\end{equation}
\end{theorem}

\begin{proof}
We do the proof by induction.

Let us consider a chain numbering of the attributes $\left\{
X_{1},X_{2},\ldots ,X_{d}\right\} $.

\begin{figure}
    \centering
    \includegraphics[width=0.3\textwidth]{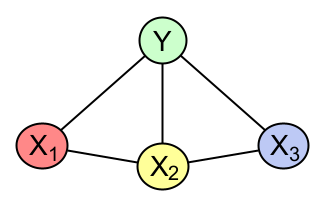}
    \caption{The initial construction of the cherry tree}
    \label{fig:formula3}
\end{figure}

First we consider the class variable $Y$ and the attributes $%
X_{1},X_{2},X_{3}$, as illustrated on Figure \ref{fig:formula3}, and we show that the formula is true for $d=3$.

As $\left( Y,X_{1},X_{2},X_{3}\right) $ is a cherry junction tree and the formula of KL divergence is given as in
(\ref{Weight_GNB}) the following formula holds 
\begin{eqnarray*}
W_{GNB}^{(3)} &=&I\left( Y,X_{1},X_{2}\right) -I\left( Y,X_{2}\right)
+I\left( Y,X_{2},X_{3}\right) = \\
&=&H\left( Y\right) +H(X_{1})+H(X_{2})-H(Y,X_{1},X_{2})+ \\
&&+H\left( Y\right) +H(X_{2})+H(X_{3})-H(Y,X_{2},X_{3})- \\
&&-\left( H\left( Y\right) +H(X_{2})-H(Y,X_{2})\right) 
\end{eqnarray*}

After reduction $W_{GNB}$ is:

\[
W_{GNB}^{(3)}=\sum%
\limits_{i=1}^{3}H(X_{i})+H(Y)-H(Y,X_{1},X_{2})-H(Y,X_{2},X_{3})+H(Y,X_{2})
\]

We add and substract $H(Y,X_{3})$ and get the following:%
\begin{eqnarray*}
W_{GNB}^{(3)}
&=&\sum%
\limits_{i=1}^{3}H(X_{i})+H(Y)+H(Y,X_{2})-H(Y,X_{1},X_{2}) \\
&&+H(Y,X_{3})-H(Y,X_{2},X_{3})-H(Y,X_{3}) \\
&=&\sum%
\limits_{i=1}^{3}H(X_{i})+H(Y)-H(X_{1}|Y,X_{2})-H(X_{2}|Y,X_{3})-H(Y,X_{3}),
\end{eqnarray*}
which is exactly the formula (\ref{keplet}) for $d=3$.

\vspace{3mm}

\begin{figure}[H]
    \centering
    \includegraphics[width=0.4\textwidth]{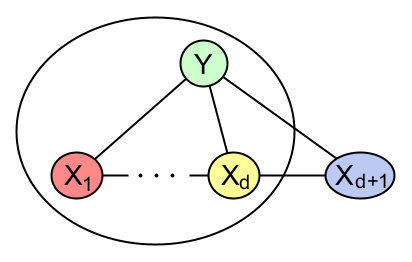}
    \caption{The inductive construction of the cherry tree in step $d$}
    \label{fig:formulad}
\end{figure}

\vspace{1mm}

Now, we suppose that the formula (\ref{keplet}) is true for $d$ attributes
and we will prove that it is true for $d+1$ attributes, see Figure \ref{fig:formulad}. Without loss of generality, we will suppose that $X_{d+1}$ is connected to $X_{d}$ and $Y$.

Since this has also a cherry-junction tree structure with a chain structure, it's
weight can be written by adding a new cluster $(Y,X_{d},X_{d+1})$  as follows:

\begin{eqnarray*}
W_{GNB}^{(d+1)} &=&W_{GNB}^{\left( d\right) }+I\left( Y,X_{d},X_{d+1}\right)
-I(Y,X_{d})= \\
&&W_{GNB}^{\left( d\right) }+H(Y)+H(X_{d})+H(X_{d+1})-H(Y,X_{d},X_{d+1}) \\
&&-\left[ H(Y)+H(X_{d})-H(Y,X_{d})\right] 
\end{eqnarray*}
After reduction we obtain:%
\[
W_{GNB}^{(d+1)}=W_{GNB}^{\left( d\right)
}+H(X_{d+1})-H(Y,X_{d},X_{d+1})+H(Y,X_{d})
\]

Now we use the induction hypothesis and get:

\begin{eqnarray*}
W_{GNB}^{(d+1)}
&=&\sum\limits_{i=1}^{d}H(X_{i})+H(Y)-\sum%
\limits_{i=1}^{d-1}H(X_{i}|Y,X_{j+1})-H(Y,X_{d}) \\
&&+H(X_{d+1})-H(Y,X_{d},X_{d+1})+H(Y,X_{d})
\end{eqnarray*}
After reduction and adding and subtracting $H(Y,X_{d+1})$ we use the
formula 
\[
H(Y,X_{d},X_{d+1})-H(Y,X_{d+1})=H\left( X_{d}|Y,X_{d+1}\right) 
\]%
and obtain

\begin{eqnarray*}
W_{GNB}^{(d+1)}
&=&\sum\limits_{i=1}^{d+1}H(X_{i})+H(Y)-\sum%
\limits_{i=1}^{d-1}H(X_{i}|Y,X_{i+1})\\
&&-H\left( X_{d}|Y,X_{d+1}\right)
-H(Y,X_{d+1})= \\
&=&\sum\limits_{i=1}^{d+1}H(X_{i})+H(Y)-\sum%
\limits_{i=1}^{d}H(X_{i}|Y,X_{i+1})-H(Y,X_{d+1})
\end{eqnarray*}

which completes the proof.
\end{proof}

\begin{theorem}
The GNB approximation gives an approximation at least as good as NB approximation.
\end{theorem}

\begin{proof}
First, let us observe that formula (\ref{Weight_GNB}) is true for any permutation of
the indices, therefore this is particularly true for the case of chain numbering.

We compare the weight of the two approximations.

\begin{eqnarray*}
W_{NB}^{(d)} &=&\sum\limits_{i=1}^{d}H\left( X_{i}\right)
+H(Y)-\sum\limits_{i=1}^{d-1}H\left( X_{i}|Y\right) -H\left( Y,X_{d}\right)
\leq  \\
&\leq
&\sum\limits_{i=1}^{d}H(X_{i})+H(Y)-\sum%
\limits_{i=1}^{d-1}H(X_{i}|Y,X_{i+1})-H(Y,X_{d})=W_{GNB}^{\left( d\right) }
\end{eqnarray*}

This inequality takes place because of  
\[
H\left( X_{i}|Y\right) \geqslant H(X_{i}|Y,X_{i+1})
\]

for all $i=1,\ldots ,d.$

The approximation with a higher weight is a better approximation, therefore
GNB gives an approximation at least as good  as Naive Bayes.
\end{proof}

\subsection{Greedy algorithm for finding a good fitting GNB structure}
\vspace{3mm}
The basic idea of this algorithm is to find an approximation with a maximum weight (\ref{Weight_GNB}).
After calculating the information of all triplets that contain the class variable $Y$ we choose the first two attributes such that the information content of $(Y,X_{i_1},X_{i_2})$ is maximum. Then we add greedily a new variable $X_{new}$  at a time, such that the maximum of 
\[
I(X_{new},X_{j},Y)-I(Y,X_{j}),
\]      
is achieved, where the variables $X_{j}$ are already
connected.
The pseudo-algorithm is given in Algorithm \ref{alg:gnba}.

\begin{algorithm}[h]
\caption{GNB-A}\label{alg:gnba}
\begin{algorithmic}
\Procedure{GNB-A}{$D$}
\State \textbf{Input:} Training samples $D$ with $d$ features $X_1, X_2, \dots, X_d$ and class variable $Y$
\State \textbf{Output:} A sorted list of triplets
\State $\mathcal{X} := \{X_1, X_2, \dots, X_d\}$
\State $X_{i_1}, X_{i_2}:=\underset{{X_i,X_j\in \mathcal{X}}}{\arg\max}\ I(Y, X_i, X_j)$
\State $L_3:=[(Y, X_{i_1}, X_{i_2})]$
\State $L_2:=[(Y, X_{i_1}), (Y, X_{i_2})]$
\State $\mathcal{X}:=\mathcal{X}\backslash\{X_{i_1}, X_{i_2}\}$
\While{$\mathcal{X}$ is not empty}
    \State $e^*, X_{new}:=\underset{{e\in L_2, X_j\in \mathcal{X}}}{\arg\max}\ (I(e, X_j)-I(e))$
    \State Append $(e^*, X_{new})$ to $L_3$
    \State Append $(Y, X_{new})$ to $L_2$
    \State $\mathcal{X}:=\mathcal{X}\backslash\{X_{new}\}$
\EndWhile
\State \textbf{return} $L_3$
\EndProcedure
\end{algorithmic}
\end{algorithm}

\vspace{1cm}

\textbf{Complexity of the GNB-A algorithm} \\

Now we will calculate the runtime of Algorithm \ref{alg:gnba}. We will use $O(f(n,d))$ to denote the total number of additions, subtractions, multiplications, divisions and comparisons for an input dataset of $n$ entries (rows) and $d$ variables $X_1,\dots,X_d$ (not including $Y$) is less than $cf(n,d)$ for some $c \in \mathbb{R}^+$.

Firstly we will show that Python 3 calculates $\log_2$ in $O(1)$ time. The Python math library uses the C programming language's built-in math package, an implementation for which can be seen in \cite{log2impl}. It uses an $8$-element Taylor series expansion for $\log(1+x)$, with a total of $29$ additions, subtractions or multiplications to calculate $\log_2(x)$ for any input $x$, meaning that it takes $29 = O(1)$ steps to calculate.

To calculate $P(X_1=x_1)$ based on a dataset with $n$ entries, we need to count how many times $X_1$ takes on $x_1$, which we will denote by $\# X_1 = x_1$, then divide by $n$:

$$P(X_1=x_1) = \frac{\# (X_1 = x_1)}{n}$$

which is $O(n)$ comparisons and $1$ division, so a total of $O(n)$ operations.

To calculate $P(X_1=x_1,X_2=x_2,X_3=x_3)$ based on a dataset with $n$ entries and at least $3$ columns, we need to evaluate a similar fraction:

$$P(X_1=x_1,X_2=x_2,X_3=x_3) = \frac{\# (X_1 = x_1, X_2 = x_2, X_3 = x_3)}{n}$$

where the nominator denotes the cases where all three equalities hold at the same time. This takes $3n$ comparisons and $1$ division, which is again an $O(n)$ amount.

To calculate the argument of the information content, we need to put all of these together:

$$\underbrace{P(x_1,x_2,x_3)}_{O(n)} \underbrace{\log_2}_{O(1)}\left(\frac{\overbrace{P(x_1,x_2,x_3)}^{O(n)}}{\underbrace{P(x_1)}_{O(n)}\underbrace{P(x_2)}_{O(n)}\underbrace{P(x_3)}_{O(n)}}\right)$$

Which takes $O(n)+O(1)+4O(n) = O(n)$ steps to evaluate.

Using this, we can show the runtime of calculating an information content $I_3$ or $I_2$ is $O(n^2)$. Let us start with $I_3$:

$$I_3(X_1,X_2,X_3) = \sum_{(x_1,x_2,x_3)} P(x_1,x_2,x_3)\log_2\left(\frac{P(x_1,x_2,x_3)}{P(x_1)P(x_2)P(x_3)}\right)$$

where the index of the summation goes through all the realizations of $(X_1,X_2,X_3)$. These triplets can be obtained by going through the dataset of $n$ rows, and storing the unique $(x_1,x_2,x_3)$ values in a dictionary, with each value of the dictionary being the number of occurrences. Such a dictionary can be constructed in $O(n^2)$ steps, since for each row of the dataset $(x_1,x_2,x_3)$, we will (potentially) need to go through the entire dictionary we have constructed so far, to see if the triplet $(x_1,x_2,x_3)$ already exists or not. If it does, then increase its value (occurrence) by one, otherwise, add it to the dictionary with value $1$.

Next, we need to sum up an $O(n)$ amount on an index set of size $O(n)$, which can also be done in $O(n^2)$ steps.

Similarly, $I_2$ can also be calculated in $O(n^2)$ steps. The difference is that the keys of the dictionary will be of size $2$ instead of $3$.

Now we can move onto the steps of Algorithm \ref{alg:gnba}.

In Step $1$, the Algorithm calculates all triplet-information contents where one of the elements is $Y$, namely

$$I_3(Y,X_i,X_j) \quad \forall i,j \in \{1,\dots,d\}$$

There are $d \choose 2$ choices for $(X_i,X_j)$, so there are $d \choose 2$ of such triplets. We have already shown that calculating an $I_3$ information content can be done in $O(n^2)$ steps, so calculating ${d \choose 2} = O(d^2)$ of them takes $O(n^2 d^2)$ operations.

In Step $2$, the Algorithm calculates all pair-information contents where one of the elements is $Y$, namely

$$I_2(Y,X_i) \quad \forall i \in \{1,\dots,d\}$$

There are $d$ options for $X_i$, so similarly to the previous case, this can be done in $O(n^2 d)$ steps.

In Step $3$, the Algorithm calculates all information content differences $I_3 - I_2$ where the arguments of $I_2$ are part of the arguments of $I_3$. These are of the form

$$I_3(Y,X_i,X_j)-I_2(Y,X_i) \quad \forall i,j \in \{1,\dots,d\}$$

and

$$I_3(Y,X_i,X_j)-I_2(Y,X_j) \quad \forall i,j \in \{1,\dots,d\}$$

So there are $2 {d \choose 2}$ of them, which is an $O(d^2)$ amount. Therefore in Step $3$, the Algorithm calculates $O(d^2)$ differences.

Finally in Step $4$, for all $k\in\{1,\dots,d-1\}$, the Algorithm chooses the $\arg\max$ of the $I_3 - I_2$ values.

In a general step $k$, the algorithm has already selected $k$ $\arg\max$'s, so there are $d-k$ that remain. In order to calculate $\arg\max (I_3 - I_2)$, we choose the highest information content nodes $(X_a,X_b)$ to add, where $X_a \in \{X_1,\dots,X_k\}$ and $X_b \in \{X_{k+1},\dots,X_d\}$. There are $k(d-k)$ such choices.

The maximum and the argmax of $k(d-k)$ elements can be obtained by going through each element once, and saving the maximum so far, and the argmax so far, which takes $k(d-k)$ comparisons.

So in steps $k=1,\dots,d-1$, the total number of comparisons is

\begin{align*}
\sum_{k=1}^{d-1} k(d-k) &= \sum_{k=1}^{d-1} kd - \sum_{k=1}^{d-1} k^2 = d\frac{(d-1)d}{2}-\frac{(d-1)d(2d-1)}{6} = \\ &= d(d-1)\left(\frac{d}{2} - \frac{2d-1}{6}\right) = d(d-1)\left(\frac{3d-2d+1}{6}\right) = \\ &= \frac{d(d-1)(d+1)}{6} = O(d^3)
\end{align*}

So overall in Steps $1$ to $4$, the Algorithm requires the following number of additions, subtractions, multiplications, divisions, or comparisons:

$$O(n^2 d^2) + O(n^2 d) + O(d^2) + O(d^3) \le O(n^2d^2 + d^3) = O(d^2(n^2+d))$$

In most cases, $d$ (the number of variables or columns) is significantly less than $n^2$ (the number of entries or rows, squared), so

$$O(d^2(n^2+d)) < O(d^2(n^2+n^2)) = O(d^2 n^2)$$

So if we have at least $\sqrt{d}$ entries in our dataset, then the runtime of the algorithm is $O(d^2 n^2)$, otherwise it is $O(d^2(n^2+d))$.

\subsection{Algorithm for finding the best fitting GNB structure}
\vspace{3mm}
\begin{algorithm}[h]
\caption{GNB-O}\label{alg:gnbo}
\begin{algorithmic}
\Procedure{GNB-O}{$D$}
\State \textbf{Input:} Training samples $D$ with $d$ features $X_1, X_2, \dots, X_d$ and class variable $Y$
\State \textbf{Output:} A list of triplets and a list of pairs.
\State $\mathcal{X} := \{X_1, X_2, \dots, X_d\}$
\State $X_{i_1}, X_{i_2}:=\underset{{X_i,X_j\in \mathcal{X}}}{\arg\max}\ I(Y, X_i, X_j)$
\State We create a $(d+1)\times (d+1)$ score matrix $S$, where the $(i,j)$-th cell is the score for the edge $j\xrightarrow{}i$: 
\begin{itemize}
    \item Row $0$ and column $0$ belongs to $Y$, row $i$ and column $i$ belongs to $X_i$.
    \item $S[i_1, 0] := I(Y, X_{i_1})$
    \item $S[i_2, i_1] := I(X_{i_1}, X_{i_2})$
    \item $S[j_1,j_2] := I(Y, X_{j_1}, X_{j_2})-I(Y,X_{j_2}),$\newline $\forall j_1 \in \{1,2,\dots,d\}\backslash\{i_1,i_2\},\ \forall j_2 \in \{1,2,\dots,d\},\ j_1\neq j_2$
    \item The remaining elements are zero.
\end{itemize}
\State We apply the Chu-Liu-Edmonds algorithm to find the maximum weighted arborescence in the directed graph belonging to $S$.
\State From the maximum weighted arborescence we create the list of triplets $L_3$ and the list of pairs $L_2$. 
\State \textbf{return} $L_3, L_2$
\EndProcedure
\end{algorithmic}
\end{algorithm}

This algorithm aims to find a GNB approximation with the maximum weight (\ref{Weight_GNB}).
After calculating the information of all triplets that contain the class variable $Y$. We choose the first two attributes where the information content of the triplet $(Y,X_i,X_j)$ is maximum. 
\begin{equation}
 (Y,X_{i_{1}},X_{i_{2}})=\underset{i,j\in V}{\arg \max }(Y,X_{i},X_{j}).
 \label{First cherry}
\end{equation}

Now we define a new directed and weighted graph on $V\cup\left\{0\right\}=\left\{ 0,1,\ldots,d\right\}$ as follows.
\begin{definition}
    We call \textbf{auxiliary directed graph structure} on the set of vertices $\left\{ 0,1,\ldots,d\right\}$, a directed graph with the following properties:
    \begin{itemize}
        \item vertex $0$ has no parent, it is called the root of the graph;
        \item the parent of vertex $i_1$ is vertex $0$;
        \item the parent of vertex $i_2$ is vertex $i_1$
        \item from vertex $i_1$ and vertex $i_2$ to all other vertices we have only one-directional edge;
        \item  for all vertices $i,j$ different from $i_1$ and $i_2$  we have an edge $(i,j)$ and an edge $(j,i)$, 
        \item there are no self-pointed edges; 
    \end{itemize}
\end{definition}

Shortly written the set of edges of the auxiliary graph structure are the following:
\begin{multline}\label{aux_edgeset}
\mathcal{E}=\left\{ \left( 0,i_1\right) \right\}\ \cup\ \left\{
(i_1,j)|j\in V\setminus\left\{i_1\right\}\right\}\
\cup\ \left\{
(i_2,j)|j\in V\setminus\left\{i_1,i_2\right\}\right\}\ \\  \cup\ \left\{(i,j)|i,j\in V\setminus\left\{i_1,i_2\right\}, i \ne j\right\}.
\end{multline}
\vspace{3mm}
Now we define the weights of the edges of the auxiliary graph.

\begin{definition}
The \textbf{weights of the edges of the auxiliary graph structure} are defined as follows:
\begin{itemize}
\item $w(0,i_1)=I(Y,X_{i_1})$
\item $w(i_1,i_2)=I(X_{i_1},X_{i_2})$
\item For any other edge $(i,j)\in \mathcal{E}\setminus\left\{ \left(
0,i_1\right) ,\left( i_1,i_2\right) \right\} $ the weight of the edge is defined
as follows:%
\begin{equation}\label{aux_edge_weight}
w\left( i,j\right) =I(Y,X_{i},X_{j})-I(Y,X_i).
\end{equation}
\end{itemize}
\end{definition}

\begin{definition}\label{aux_graph}
The weighted graph $G(V,\mathcal{E})$, where the set of edges $\mathcal{E}$ is defined by (\ref{aux_edgeset}) and
the weights of the edges are defined in (\ref{aux_edge_weight}) is called \textbf{auxiliary graph of the GNB graph}.
\end{definition}

\bigskip Now we recall the definition of an arborescence in a directed graph.

\begin{definition} \label{arborescence}
An \textbf{arborescence} is a directed graph with a root vertex $0$ and with all other vertices $v$ having the property that there is exactly one directed path from $0$ to $v$.
\end{definition}

There are some important equivalent definitions. An arborescence can be defined as a rooted digraph in which the path from the root to any other vertex is unique.

\bigskip 

For finding the maximum weighted arborescence in a directed graph, Chu Liu Edmond's algorithm \cite{edmonds1967optimum} was proposed.

\begin{remark}
\label{edges in aux}
The maximum arborescence of the auxiliary graph contains the edges $(0,i_1)$
and $(i_1,i_2)$.    
\end{remark}
\begin{proof}
This is true because of Definition \ref{arborescence} and since vertex $i_1$'s
single parent is vertex $0$, and vertex $i_2$'s single parent is vertex $i_1$.
\end{proof}

\begin{theorem}
Each vertex of an arborescence has only one parent.
\end{theorem}

\begin{proof}
If a vertex $v$ would have two parents $p_{1}(v)$ and $p_{2}(v)$ then by definition for each of the parents we have a single directed path from the root, then it is obvious that we would have two directed paths starting from the root which lead to $v$, which contradicts Definition \ref{arborescence}.
\end{proof}

We apply Chu Liu Edmonds algorithm \cite{edmonds1967optimum} to the auxiliary graph.

The algorithm's output is a maximum weighted arborescence with the property that for each vertex $ i\in V$ a unique directed path from $0$ to $i$ is defined.

\begin{definition}\label{Optimized Generalized Naive Bayes}
We define the \textbf{Optimized Generalized Naive Bayes} structure based on the maximal weighted arborescence as follows:

\begin{itemize}
\item The triplet $(Y,X_{i_1},X_{i_2})$ defined by formula (\ref{First cherry}) is the first cluster of the cherry tree.

\item For any $k\in V\setminus\left\{i_1,i_2\right\}$ there exists a unique directed path $0,i_1,i_{k_{1}}\ldots i_{k_{s}}=k$ from $0$ to $k$, such that $i_{k_1},\ldots i_{k_{s}}\in V$.
Based on this we will define the other clusters of the cherry tree as follows $%
(Y,X_{p(k_{i})},X_{k_{i}})$, for all $k_{i}\notin \left\{ 0,i_1,i_2\right\} $, $k_{i}\in
\{i_{k_{1}},\ldots i_{k_{s}}=k\}$
\end{itemize}
\end{definition}

Now we can state an important result of our paper.

\begin{theorem}
The Optimized Generalized Naive Bayes given in Definition \ref{Optimized Generalized Naive Bayes} is the
maximum weighted Generalized Naive Bayes distribution if this structure contains the maximum weighted cherry $(Y, X_{i_1}, X_{i_2})$.

\begin{proof}
From Remark \ref{edges in aux} we have that the vertices $0,i_1,i_2$ will be in the tree this way the weight of the approximation will contain $%
I(Y,X_{i_1},X_{i_2})$. 

Any \ cluster $(Y,X_{p(k_{s})},X_{k_{s}})$ is connected to the already
connected vertex by the separator $(Y,X_{p(k_{s})})$, and  adds to the weight of approximation the term $I(Y,X_{p(k_{s})},X_{k_{s}})-I(Y,X_{p(k_{s})})$, the weight of the corresponding edge in the auxiliary graph.

If we would have a GNB with a larger weight, then the maximum weighted arborescence would be not optimal, which is in contradiction with Chu Liu Edmonds algorithm, which guarantees to find the maximum weighted arborescence.
\end{proof}
\end{theorem}

A natural question is how to get the Weight of the Optimized GNB approximation from the weight of the maximum weighted arborescence (output of Chu-Liu-Edmonds algorithm). 

The weight of the arborescence differs from the weight of the optimal approximation by the weight of the first two edges, which are present in the weight of the arborescence, and are not present in the weight of GNB, therefore we have to subtract them. Although the weight of the GNB approximation contains the information content of the first triplet, which is not contained by the weight of arborescence. This is why we have to add it to it.

\vspace{4mm}

\textbf{Complexity of the GNB-O algorithm} \\

The GNB-O algorithm first fills out the $S$ matrix by calculating all pair and triplet information contents, both of which we have seen can be done in $O(n^2d^2)$ time. Then we run the Chu Liu Edmonds algorithm on the directed graph belonging to $S$. Its runtime is $O(|E|\cdot|V|)$, where $|E|$ are the number of edges and $|V|$ are the number of vertices. In our case, $|V|=d+1=O(d)$ and $|E|=(d-2)^2+2=O(d^2)$, so the runtime of the Chu-Liu-Edmons algorithm is
$$O(|E|\cdot|V|)=O(d^2\cdot d)=O(d^3)$$
Once again, similarly to GNB-A, the total runtime of GNB-O is
$$O(n^2d^2+d^3)=O(d^2(n^2+d))\stackrel{(*)}{=}O(n^2d^2)$$
where $(*)$ holds true if $d<n^2$, so the number of attributes (columns) is less than the number of data (rows) squared; which is true in almost all cases.

\subsection{On how the introduced GNB concepts are related to former improvements of NB}
The most closely related structures and algorithms are the TAN, ATAN, and Ghofrani algorithms, and they are given in \cite{friedman1997bayesian} and \cite{ghofrani2018new}, while their complexities are shown in \cite{ghofrani2018new} (Page 6). This part is dedicated to discussing the similarities and differences between structures, algorithms, and complexities. 
From a structural point of view, we proved in Theorem \ref{tree between X}, that the graph structure defined on the explanatory variables is a tree, which is also the case in the graph structures used by TAN and ATAN algorithms.

The time complexity of the TAN algorithm is
$$O(nd^2)+O(k(dv)^2)+O(d^2\log(d))$$
While the time complexity  of the ATAN algorithm it is $$O(nd^2)+O(k(dv)^2)+O(d^3\log(d)),$$
where $n$ is the number of training data (rows), $d$ is the number of features (columns), $v$ is the average number of values per feature (which we have set to $5$ in our evaluation), and $k$ is the number of classes (which in our evaluation setting is $2$, a binary $0-1$ value).

The time complexity of the Ghofrani algorithm is
$$O(nd^{l_c})+O(k(dv)^2)+O(d^2\log(d))+O(d^3l_c^3)+O(nkn_{clq})$$
where $n,d,v,k$ are the same as before, $l_c$ is the dimensionality of the table containing probability estimates for each class (in our case, $2$), and $n_{clq}$ is the number of cliques (clusters) generated by the algorithm (which in our case is $d$).

To be able to compare the runtime of our algorithm with TAN, ATAN, and Ghofrani's algorithm, we will use the substitutions $v=5$, $k=2$, $l_c=2$ and $n_{cql}=d$;
the runtime of TAN is therefore
$$O(nd^2)+O(d^2)+O(d^2\log(d)) \le O(d^2(n+\log(d)))$$

The runtime of ATAN is therefore

$$O(nd^2)+O(d^2)+O(d^3\log(d)) \le O(d^2(n+d\log(d)))$$

The runtime of Ghofrani's algorithm is therefore
\begin{align*}O(nd^2)+O(d^2)+O(d^2\log(d))+O(d^3)+O(nd) \le O(nd^2+d^3) = \\ = O(d^2(n+d))\stackrel{(*)}{\le} O(d^2\cdot 2n)=O(d^2n)\end{align*}

where $(*)$ holds true if there is more training data (rows) than attributes (columns), which is usually the case.

We have already seen that the runtime of GNB-A is $O(d^2n^2)$, which means that TAN is faster as we can assume that $\log(d) < n$, while ATAN is also slightly faster if $n+d\log(d) < n^2$, which holds true for most datasets (which have more training data than features), since
$$n+d\log(d) < n+d\frac{d}{2} < n+n\frac{n}{2} = n+\frac{n^2}{2} < n^2$$
Furthermore, Ghofrani's LDMLCS is also faster if we set $l_c=2$, as its runtime is $O(d^2n)$.

The novelty of the LDMLCS algorithm introduced by Ghofrani et.al.in \cite{ghofrani2018new} was that it exploits the concept of conditional independence between random variables in the context of probabilistic graphical models. The cherry tree-based algorithms, the GNB-A and GNB-O algorithms exploit conditional independence to maximize the information content of the model and to reduce redundancy. 

The main difference, from an information-theoretical point of view, between LDMLCS algorithm in \cite{ghofrani2018new}, the ATAN in \cite{jiang2012improving}  and the algorithms introduced here is that the LDMLCS and ATAN consider only the bivariate conditional mutual information to weigh the edges between the explanatory attributes whereas, in our algorithms, we use the fact that any structure that has a tree-like structure between the explanatory variables has a special cherry tree probability distribution. Based on this the expression of Kullback Leibler divergence (\ref{KL_GNB_th}) between this probability distribution and the sample data depends on the graph structure. By minimizing the KL divergence we find the graph structure.
We emphasize here that in Ghofrani's paper \cite{ghofrani2018new}, as an alternative method they apply the cherry tree algorithm to the set of attributes, which is different from our methods introduced here, where each cherry contains the classifier.

\section{The classification process and feature selection methods}
\vspace{3mm}
Let us suppose that based on the training data we fitted a GNB probability distribution, by GNB-A or GNB-O. We get as output the set of clusters and the set of separators of the GNB. Now we present the main steps of the classification process.

\begin{description}
\item[Input:] The set of clusters and the set of separators which contain the classification variable $Y$:
\begin{itemize}
\item $\mathcal{C}\left( Y\right) =\left\{ C_{k}\left( Y\right)=\left( Y,X_{i_{k}},X_{j_{k}}\right) ~|~\left(
Y,X_{i_{k}},X_{j_{k}}\right) \in \mathcal{C}\right\} $ and the
corresponding marginal probability distributions,

\item $\mathcal{S}\left( Y\right) =\left\{ S_{k}\left(Y\right)
=\left(Y,X_{i_{k}}\right) ~|~\left( Y,X_{i_{k}}\right) \in \mathcal{%
S}\right\} $ and the corresponding marginal probability distributions.
\end{itemize}

\item[Output:] A probability distribution over the classes $y\in \left\{1,\ldots ,m\right\} $ and the classification of the element into the most probable class.

\item[Step 1:] For $y=1,\dots,s$ calculate the approximation of the probability distribution of the training data by using the cherry tree approximation:

\begin{multline}\label{eq:1}
\widetilde{P}\left( Y=y,X_{1}=x_{1},\ldots ,X_{d}=x_{d}\right) =\\
=\frac{\prod\limits_{\left( Y,X_{i_{k}},X_{j_{k}}\right) \in \mathcal{C}}P\left( Y=y,X_{i_{k}}=x_{i_{k}},X_{j_{k}}=x_{j_{k}}\right) }{\prod\limits_{\left( Y,X_{i_{k}}\right) \in \mathcal{S}}P\left(
Y=y,X_{i_{k}}=x_{i_{k}}\right) ^{v_{k-1}}}.
\end{multline}

\item[Step 2:] 
\begin{itemize}
    \item If there exist $y\in \left\{ 1,\ldots ,s\right\} $ such that 
    \[
    \widetilde{P}\left( Y=y,X_{1}=x_{1},\ldots ,X_{d}=x_{d}\right)\neq 0,
    \] 
    then the vector $(X_{1}=x_{1},\ldots ,X_{d}=x_{d})$ is classified in the class $y$ with probability 
    \begin{multline*}
        \widetilde{P}\left( Y=y|X_{1}=x_{1},\ldots ,X_{d}=x_{d}\right) =\\
        =\frac{\widetilde{P}\left( Y=y,X_{1}=x_{1},\ldots ,X_{d}=x_{d}\right) }{\sum\limits_{y=1}^{s}\widetilde{P}\left( Y=y,X_{1}=x_{1},\ldots,X_{d}=x_{d}\right)},
    \end{multline*}
    \begin{equation*}
    y^{\ast }=\underset{y}{\arg \max }\widetilde{P}\left(Y=y,X_{1}=x_{1},\ldots ,X_{d}=x_{d}\right) ,~y\in \left\{ 1,\ldots,s\right\}.
    \end{equation*}
\item Else, go to Step 3.
\end{itemize}

\item[Step 3:]For all  $k$ with  $P\left(Y=y,X_{i_{k}}=x_{i_{k}},X_{j_{k}}=x_{j_{k}}\right) =0$ calculate \[ P\left( Y=y,X_{i_{k}}=x_{i_{k}}\right)\textrm{ and }P\left( Y=y,X_{j_{k}}=x_{j_{k}}\right)\]
\begin{itemize}
    \item If for any $i_k$: $P\left(Y=y,X_{i_{k}}=x_{i_{k}}\right)=0$, substitute it with respect to
    \begin{equation}\label{eq:2}
        P\left( Y=y,X_{i_{k}}=x_{i_{k}}\right):=P\left( Y=y\right)\cdot P\left(X_{i_{k}}=x_{i_{k}}\right),
    \end{equation}
    then go to Step 4.
    \item Else, go to Step 4.
\end{itemize}

\item[Step 4:] Calculate
\begin{multline}\label{eq:3}
P\left(Y=y,X_{i_{k}}=x_{i_{k}},X_{j_{k}}=x_{j_{k}}\right) :=\\
\frac{P\left( Y=y,X_{i_{k}}=x_{i_{k}}\right) P\left(Y=y,X_{j_{k}}=x_{j_{k}}\right) }{P\left( Y=y\right) }
\end{multline}
Go to Step 1.
\end{description}

\subsection{Feature selection, feature importance score}
\vspace{3mm}
The feature selection can be achieved in two stages. The first stage is during the learning process. The relevant features can be chosen based on how the information added by joining a variable increases the weight (\ref{Weight_GNB}) of the constructed cherry tree.
We have to highlight here, that the importance score of an attribute $X_{i_k}$ will express the amount of information added by it, taking into account the information added by all $X_{i_1}$\dots $X_{i_{k-1}}$ attributes.

Feature selection can be achieved also in the second stage, based on the validation set. In this case, the users/ experts can decide on relevant features (how many triplets respectively attributes they should use for the classification) based on the graphs for precision, recall, f1 -score, and AUC-score. 
We have to underscore here, that the role of a new attribute $X_{i_k}$ added by a new triplet is mirrored by how the given accuracy measure changes by adding the attribute $X_{i_k}$to the $X_{i_1}$\dots $X_{i_{k-1}}$  attributes already added to the model.

In the following, we discuss the feature selection, and feature scores for both stages, for the two algorithms introduced separately. 
\vspace{3mm} 

 \textbf{Feature selection in stage 1 in the case of Algorithm GNB-A} 
 
 In this case, in each step, the weight of the cherry tree is increased by adding the new triplet's information content and subtracting the bi-variate separator's information content (two-node edge) to which the new attribute is connected. This way after each step, the structure remains a GNB on the given set of explanatory variables. 
This method chooses the features in every step in a greedy way. The algorithm maximizes added information and minimizes redundancy at the same time, in each step.
Based on the graph wich shows how the added variables influence the cumulation of the information it is straightforward to decide on which explanatory variable we chose.
\vspace{3mm} 

 \textbf{Feature selection in stage 2 in the case of Algorithm GNB-A}. 
 
In this case, to the ordered triplets added in each step (in fact by adding a new feature), the following accuracy scores are assigned: precision-, recall-, AUC and f1 -score. Based on these scores the expert can choose which and how many explanatory variables are needed for a given purpose. 

\vspace{3mm} 

 \textbf{Feature selection in stage 1 in the case of Algorithm GNB-O}.
 
In the case of Algorithm \ref{alg:gnbo} is not straightforward to obtain the ordered clusters of the cherry tree.
First, let us recall here, that if the triplet with the largest weight (information content) is contained in the optimized GNB, then the GNB-O Algorithm gives the best-fitting GNB, taking into consideration all the attributes. 
The output of the algorithm is the set of all triplets without an ordering of them. To assign an importance score to the attributes first we have to order them such that any ordered subset of successive clusters has a cherry tree structure.

The chain numbering of the attributes is a very important concept here. In this case the attributes $X_1$,\dots,$X_d$, are reordered $X_{i_1}$,\dots,$X_{i_d}$, by a permutation. This ordering is essential, when we assign to the attributes $X_{i_3}$,\dots,$X_{i_d}$ an importance score. 

Finding this ordering is also at the heart of the accuracy measure representations on the validation set, with regard to how many attributes were taken into consideration.

\vspace{3mm}

Now we will give a method to get the chain ordering from the output tree representation of the Chu-Liu-Edmonds Algorithm.

We build a score matrix which contains on $(i,j)$ position the weight of the $(i,j)$ directed edge. 
The main idea of the numbering algorithm is to find in each step the "leaf vertices" i.e. those vertices that are not parents, and choose the one with the smallest weight, then delete it from the columns and rows.
The order of deleting of the vertices gives the descending ordering of attributes. 

\textbf{Feature selection in stage 2 in the case of Algorithm GNB-O}.

In this case, to the ordered triplets (based on the chain ordering procedure), we assign respectively precision-, recall-, and f1 -scores. These will be the feature importance scores, based on these scores the expert can choose which and how many explanatory variables are needed for the given purpose. 
\vspace{3mm}
In the following section, we present how our new methods work on different datasets, and how can we use the feature scores obtained in the second stage.

\section{Numerical Results}

The introduced algorithms are designed for the classification task. We will use evaluation metrics specific to this problem. 
\subsection{Evaluation metrics}
In binary classification, beyond accuracy, precision and recall are often used metrics to evaluate the models. Let the labels we want to predict be `positive' and `negative', then we can define the \emph{confusion matrix}, in which the columns show the actual values and the rows show the predicted values:
\begin{table}[h!]
\begin{tabular}{cc|c|c|}
\cline{3-4}
\multicolumn{2}{c}{\multirow{2}{*}{}} & \multicolumn{2}{|c|}{Actual} \\ \cline{3-4} 
\multicolumn{2}{c|}{} & Positive & Negative \\ \hline
\multicolumn{1}{|c|}{\multirow{2}{*}{Predicted}} & Positive & \#\{\textbf{True Positive}\} & \#\{\textbf{False Positive}\} \\ \cline{2-4} 
\multicolumn{1}{|c|}{} & Negative & \#\{\textbf{False Negative}\} & \#\{\textbf{True Negative}\} \\ \hline
\end{tabular}
\end{table}

There are four possible outcomes according to the actual and the predicted label. The elements of the matrix show the occurrence of each outcome. Based on these accuracy, precision and recall can be defined as follows:

\begin{equation*}
    \textrm{Accuracy} = \frac{\#\{\textrm{True Positive}\}+\#\{\textrm{True Negative}\}}{\#\{Total\}}
\end{equation*}

\begin{equation*}
    \textrm{Precision} = \frac{\#\{\textrm{True Positive}\}}{\#\{\textrm{True Positive}\}+\#\{\textrm{False Positive}\}}
\end{equation*}

\begin{equation*}
    \textrm{Recall} = \frac{\#\{\textrm{True Positive}\}}{\#\{\textrm{True Positive}\}+\#\{\textrm{False Negative}\}}
\end{equation*}

%\begin{equation*}
%    \textrm{F1 score} = \frac{2\cdot\#\{\textrm{True Positive}\}}{2\cdot\#\{\textrm{True Positive}\}+\#\{\textrm{False Positive}\}+\#\{\textrm{False Negative}\}}
%\end{equation*}

\begin{equation*}
    \textrm{F1 score} = 2\times \frac{\textrm{Precision}\times \textrm{Recall}}{\textrm{Precision}+\textrm{Recall}}
\end{equation*}

\subsection{Data preparation}\label{dataprep}
The algorithms introduced in this paper are developed for discrete data. In general, the datasets contain discrete, categorical, and continuous attributes. The categorical variables can be treated similarly to the discrete ones. The algorithms introduced in this paper are based on the information content, which does not depend on the "values" taken on by the variables, only on the probabilities. This way the algorithm is more flexible in dealing with different kinds of attributes (categorical, discrete).
The continuous attributes are discretized. There are many possibilities for discretization. We chose here a quite general method, which can be applied to different types of datasets based on quantiles. However, there are many interesting new discretization methods, even based on Naive Bayes, such as \cite{zhang2023rigorous} and \cite{wang2024max}. 

The quantile discretization which we use here will be described shortly in the following.
If a variable $X_i$ takes on more than $5$ distinct values, then it is discretized into at most $5$ values. Denoting the individual values of $X_i$ by $X_i = [x_1,x_2,\dots,x_n]$, let us first make an ordered list $\hat{X}_i := [x_{\phi(1)},\dots,x_{\phi(n)}]$ with $x_{\phi(j)} \le x_{\phi(j+1)}\ \forall{j}\in\{1,\dots,n-1\}$. Then the quantiles of this attribute are

$$Q_j = x_{\lfloor\phi(nj/5)\rfloor} $$

For $j \in \{1,\dots,4\}$, in our case ($4$ quantiles along with the $0\%$-quantile and the $100\%$-quantile split the interval into $5$ parts). If for any $i \in \{1,\dots,5\}$, $Q_i = Q_{i+1}$, then we threw away $Q_{i+1}$, and only used $Q_i$. Let us denote the number of quantiles remaining after this process by $M \le 5$. And let $Q_0 = \min X_i$, $Q_{M+1} = \max X_i$.

Then we let the average of all discretized intervals be the representative value for each interval:

$$a_j = \frac{1}{|\{i: Q_j \le x_i < Q_{j+1}\}|}\sum_{i=0}^n x_i \textbf{1}(Q_j \le x_i < Q_{j+1}) \quad \forall j \in \{1,\dots,M\}$$

Where $\textbf{1}$ is the indicator function. Finally, the discretized elements of $X_i = [x_1,x_2,\dots,x_n]$ are $D_i = [d_1,d_2,\dots,d_n]$, where

$$d_k = \sum_{s=0}^M a_s \textbf{1}(Q_s \le x_k < Q_{s+1}) \quad \forall k \in \{1,\dots,n\}$$

\subsection{Description of Datasets}
%Detection of breast cancer is crucial for successful treatment. Conventional methods such as breast biopsy are invasive and must be performed by experts. However, samples obtained with less invasive techniques like fine needle aspiration can be easily digitized and used for computer-aided diagnosis. We illustrate here the algorithms on the Diagnostic Wisconsin Breast Cancer Database available at \url{http://archive.ics.uci.edu/ml/machine-learning-databases/breast-cancer-wisconsin/}. The aim is to classify a tumor as benign or malignant. The data is slightly unbalanced 0.63 of the data are benign (B) and 0.37 of the data are malign (M). It is important to notice that between some of the explanatory variables (attributes), there is a strong correlation.

\begin{center}
\begin{tabular}{ |c|c|c|c|c| }
 \hline
 & Num. of & Num. of & Num. of & Num. of \\ 
 Dataset & attributes & continuous & discrete & entries  \\ 
  & (columns) & attributes & attributes & (rows) \\ 
 \hline
 Wdbc [ref] & 30 & 30 & 0 & 569 \\
 Heart Disease [ref] & 13 & 5 & 8 & 297 \\
 Diabetes [ref] & 16 & 1 & 15 & 519 \\  
 Thyroid Ann Train [ref] & 21 & 15 & 6 & 3771 \\
 Parkinson [ref] & 22 & 22 & 0 & 197\\
 \hline
\end{tabular}
\end{center}
In all datasets we discretized the continuous attributes with the method explained in Subsection~\ref{dataprep} and omitted all rows with missing values. 
It is important to underscore here that the accuracy results usually may depend on the discretization methodology. In real-life problems experts may have a better insight into the data, therefore they can use a better method to discretize the continuous variables by choosing meaningful intervals. To have a more general approach here, and to be able to apply the same methodology to all datasets, we chose the quantile discretization.

All of our results are based on the same type of discretization, therefore these results should be not compared with other results based on other discretization. The aim of the paper was to present the GNB structure and to compare it with other algorithms by using a general discretization method.

The algorithms were run $5$ times on each dataset with a random $15\%$ test set selection, using various numbers of triplets. The average results of these runs can be seen in this subsection (Figure \ref{fig:resultsWdbc}, \ref{fig:resultsHeartDisease}, \ref{fig:resultsDiabetes}, \ref{fig:resultsThyroidAnnTrain}).

\subsubsection{Wdbc dataset}
\begin{figure}[H]
    \centering
    \begin{subfigure}[b]{0.45\textwidth}
        \centering
        \includegraphics[width=0.95\textwidth]{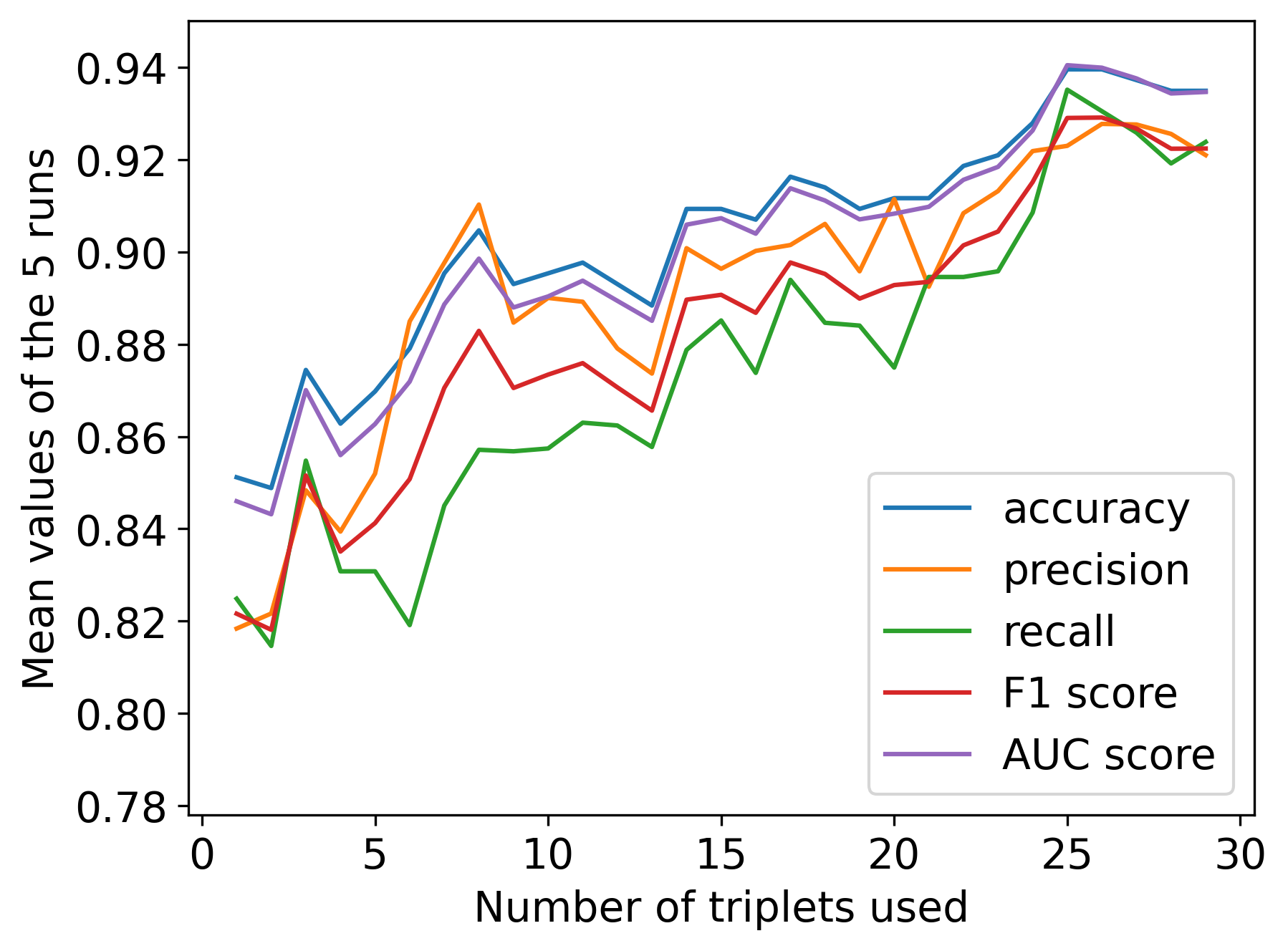}
        \caption{Results of the GNB-A algorithm on the Wdbc dataset}
        \label{subfig:wbdc_gnba}
    \end{subfigure}
    \begin{subfigure}[b]{0.45\textwidth}
        \centering
        \includegraphics[width=0.95\textwidth]{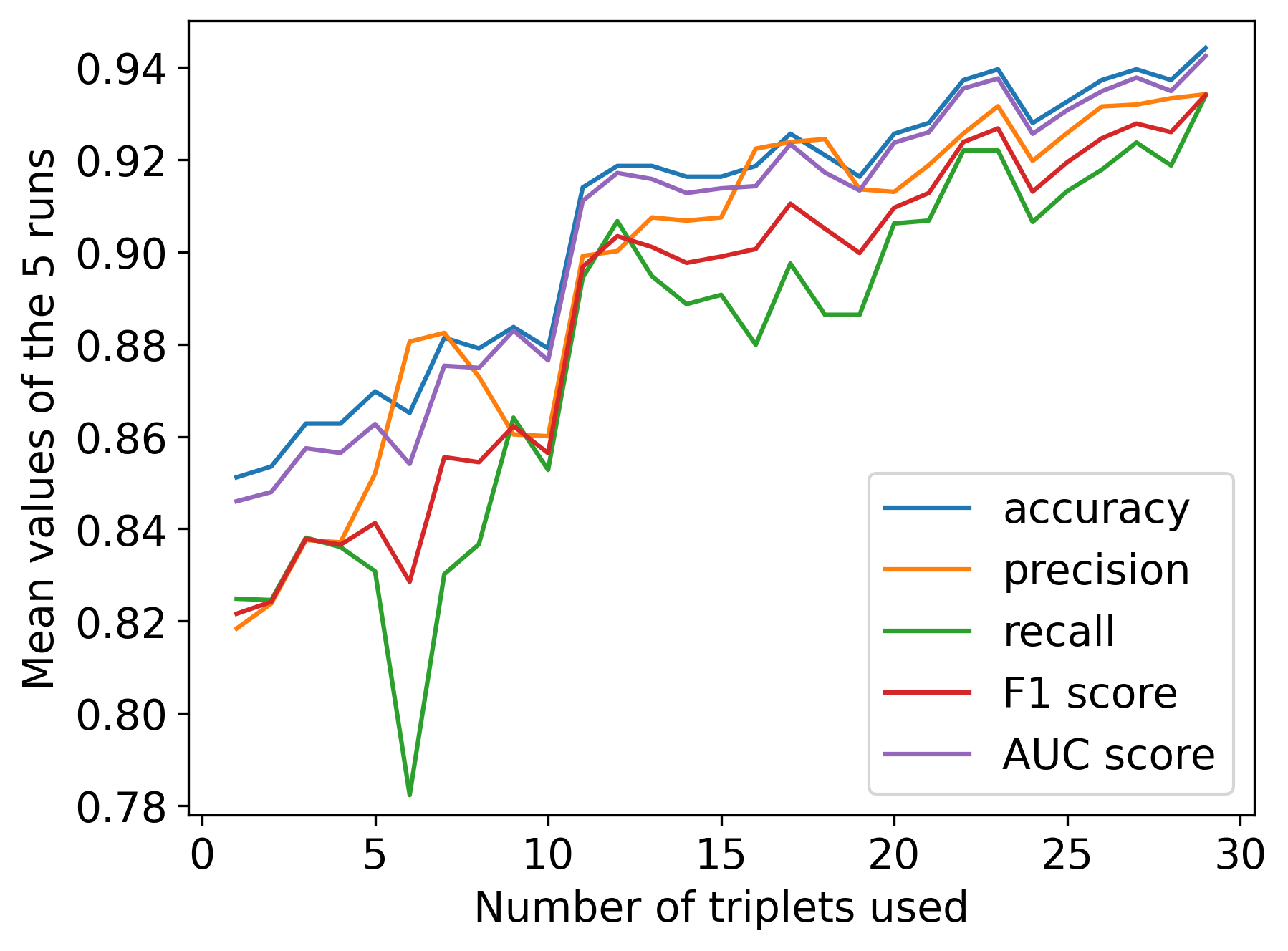}
        \caption{Results of the GNB-O algorithm on the Wdbc dataset}
        \label{subfig:wbdc_gnbo}
    \end{subfigure}
    \caption{}
    \label{fig:resultsWdbc}
\end{figure}

In the Diagnostic Wisconsin Breast Cancer Database (\url{https://archive.ics.uci.edu/dataset/17/breast+cancer+wisconsin+diagnostic}) the aim is to classify a tumor as benign or malignant. The data is slightly unbalanced 0.63 of the data are benign (B) and 0.37 of the data are malign (M).
The average results of the runs can be seen in Figure \ref{fig:resultsWdbc}.
In Figure \ref{fig:resultsWdbc} of GNB-A one can observe that the highest score is achieved with 25 variables whereas in the case of GNB-O the scores typically improve until it uses all 30 explanatory variables. It is also interesting to remark, that by using these methods on this dataset over-fitting is avoided.

\subsubsection{Heart Disease dataset}
\begin{figure}[H]
    \centering
    \begin{subfigure}[b]{0.45\textwidth}
        \centering
        \includegraphics[width=0.95\textwidth]{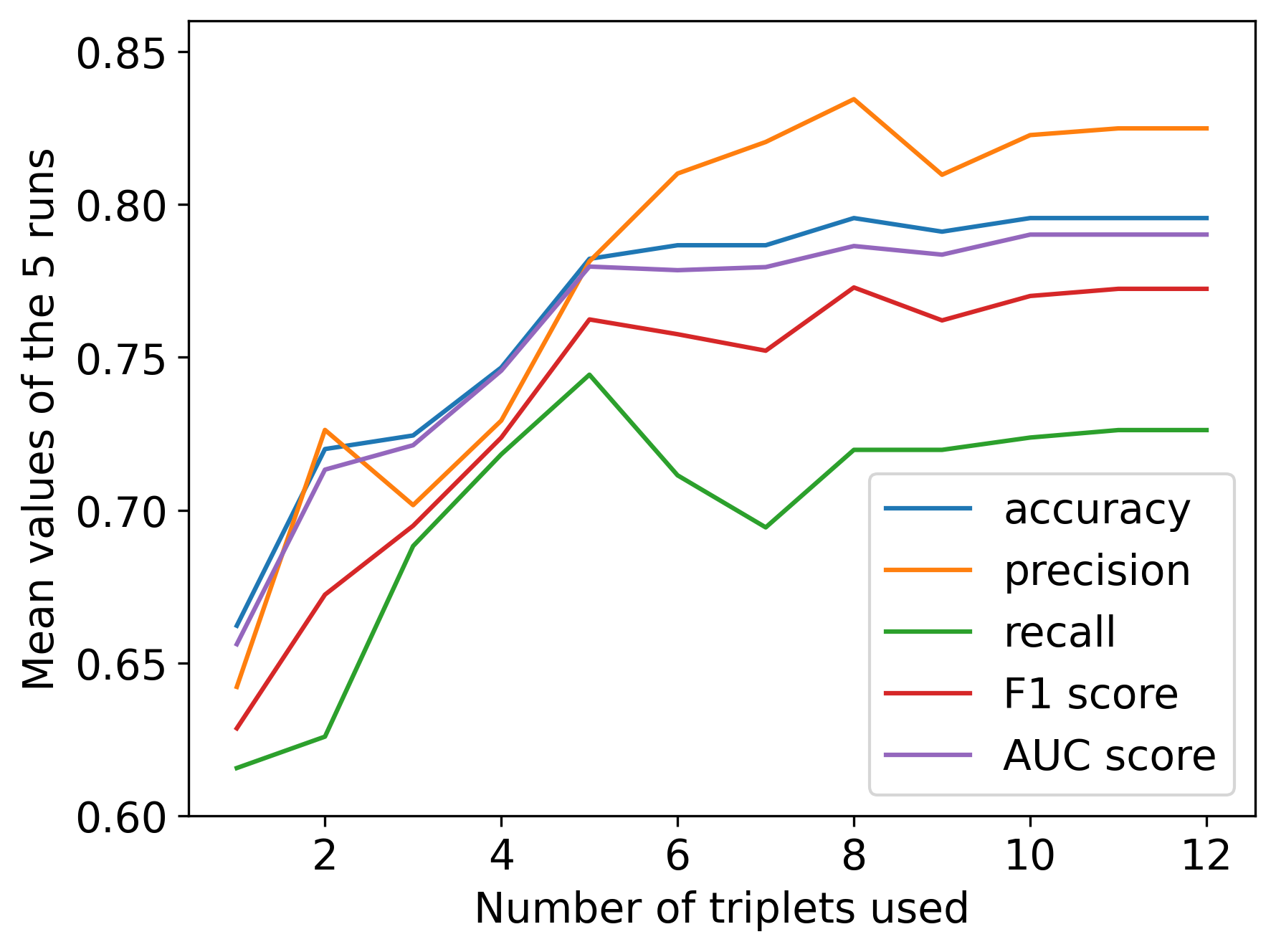}
        \caption{Results of the GNB-A algorithm on the Heart Disease dataset}
        \label{subfig:hd_gnba}
    \end{subfigure}
    \begin{subfigure}[b]{0.45\textwidth}
        \centering
        \includegraphics[width=0.95\textwidth]{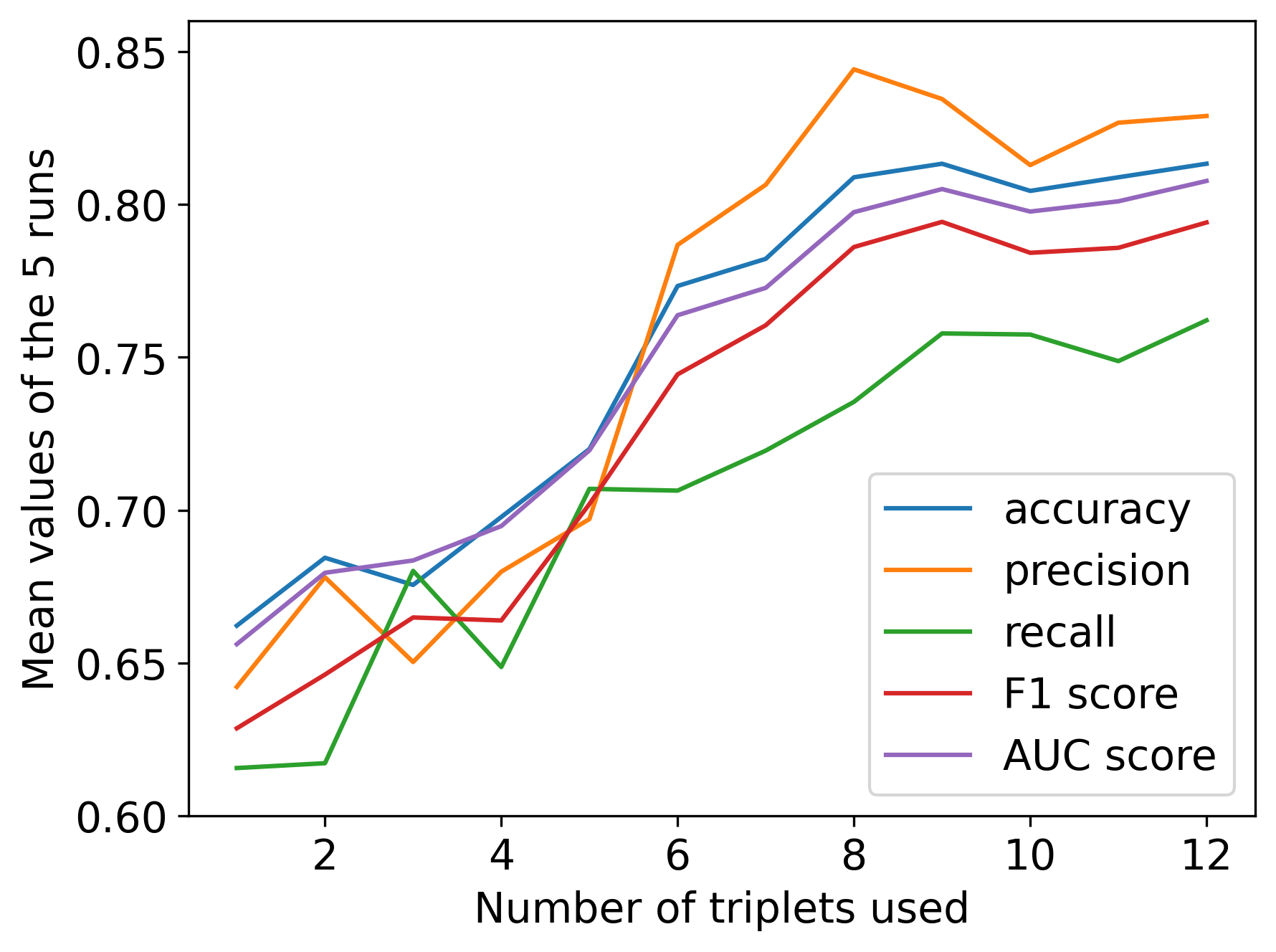}
        \caption{Results of the GNB-O algorithm on the Heart Disease dataset}
        \label{subfig:hd_gnbo}
    \end{subfigure}
    \caption{}
    \label{fig:resultsHeartDisease}
\end{figure}

The Heart Disease database (\url{https://archive.ics.uci.edu/dataset/45/heart+disease}) contains 4 datasets from which we used the Cleveland dataset as it contained the fewest missing values, 6 rows in all which were left. The class variable takes on integer values from 0 (no presence of heart disease) to 4. Still, we concentrated on simply distinguishing the presence (values 1, 2, 3, 4) from the absence (value 0) of the disease in a patient, so we joined classes 1, 2, 3 and 4 into one class.
The average results of the runs can be seen on Figure \ref{fig:resultsHeartDisease}.
One can observe in Figure \ref{fig:resultsHeartDisease} that 8 explanatory variables seem to give the best scores, which cannot be improved further. 

\subsubsection{Diabetes dataset}

\begin{figure}[H]
    \centering
    \begin{subfigure}[b]{0.45\textwidth}
        \centering
        \includegraphics[width=0.95\textwidth]{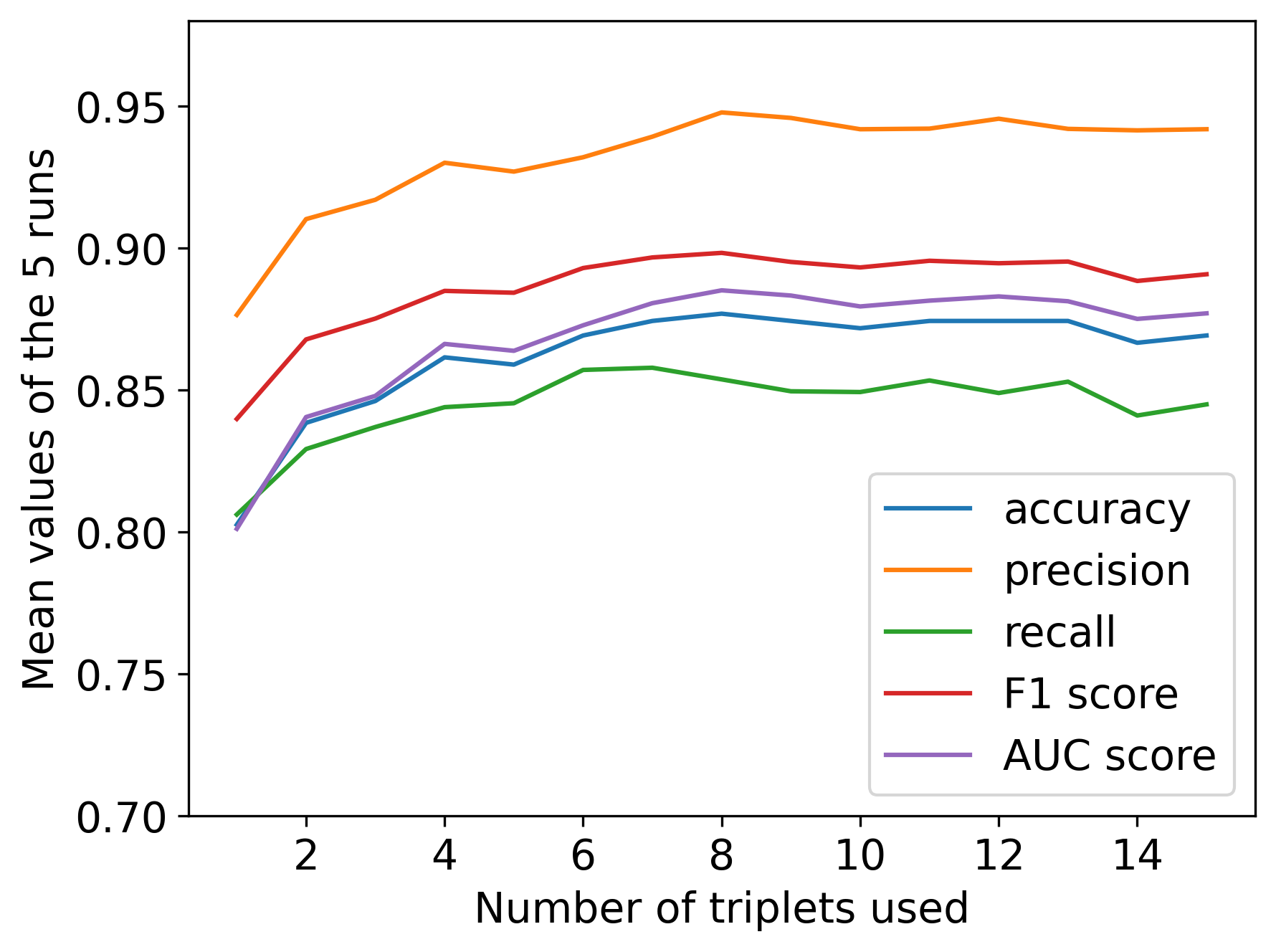}
        \caption{Results of the GNB-A algorithm on the Diabetes dataset}
        \label{subfig:diabetes_gnba}
    \end{subfigure}
    \begin{subfigure}[b]{0.45\textwidth}
        \centering
        \includegraphics[width=0.95\textwidth]{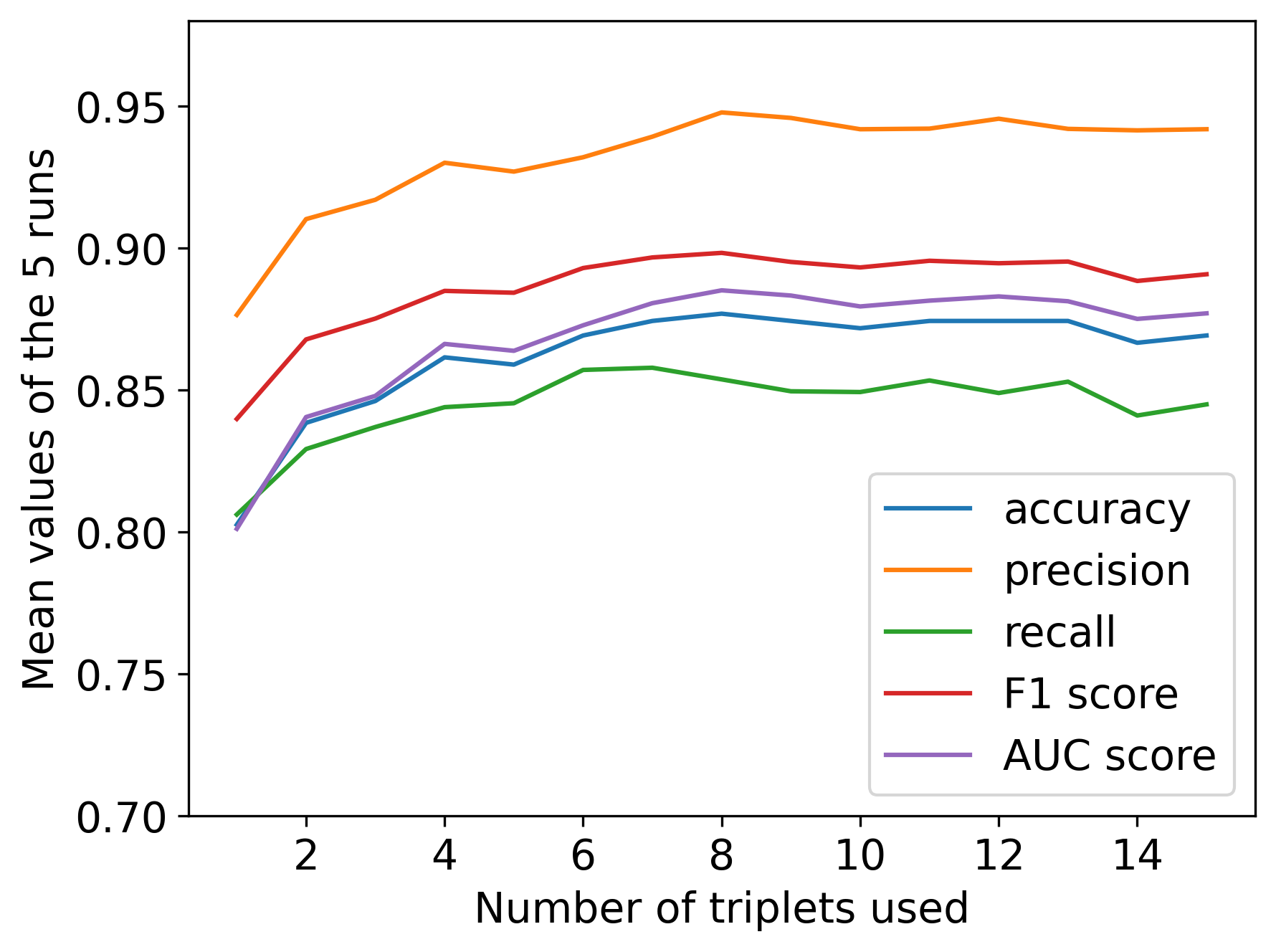}
        \caption{Results of the GNB-O algorithm on the Diabetes dataset}
        \label{subfig:diabetes_gnbo}
    \end{subfigure}
    \caption{}
    \label{fig:resultsDiabetes}
\end{figure}

This dataset contained no missing values and only one continuous attribute (Age), which was discretized with the method explained in Subsection~\ref{dataprep}. The algorithm was ran $5$ times on it with a random $15\%$ test set selection. The average results of the runs can be seen on Figure \ref{fig:resultsDiabetes}.

It seems that if a small number of triplets are desired, then using around $8$ of them will result in a high accuracy, precision, recall, and F1 score; These scores remain relatively large by further adding new triplets. 

\subsubsection{Thyroid dataset}

The Thyroid dataset contained multiple different data frames, of which we have selected the Thyroid Ann Train dataset, as it contained mostly binary values and enough rows to reliably calculate the information contents of the selected triplets.

\begin{figure}[H]
    \centering
    \begin{subfigure}[b]{0.45\textwidth}
        \centering
        \includegraphics[width=0.95\textwidth]{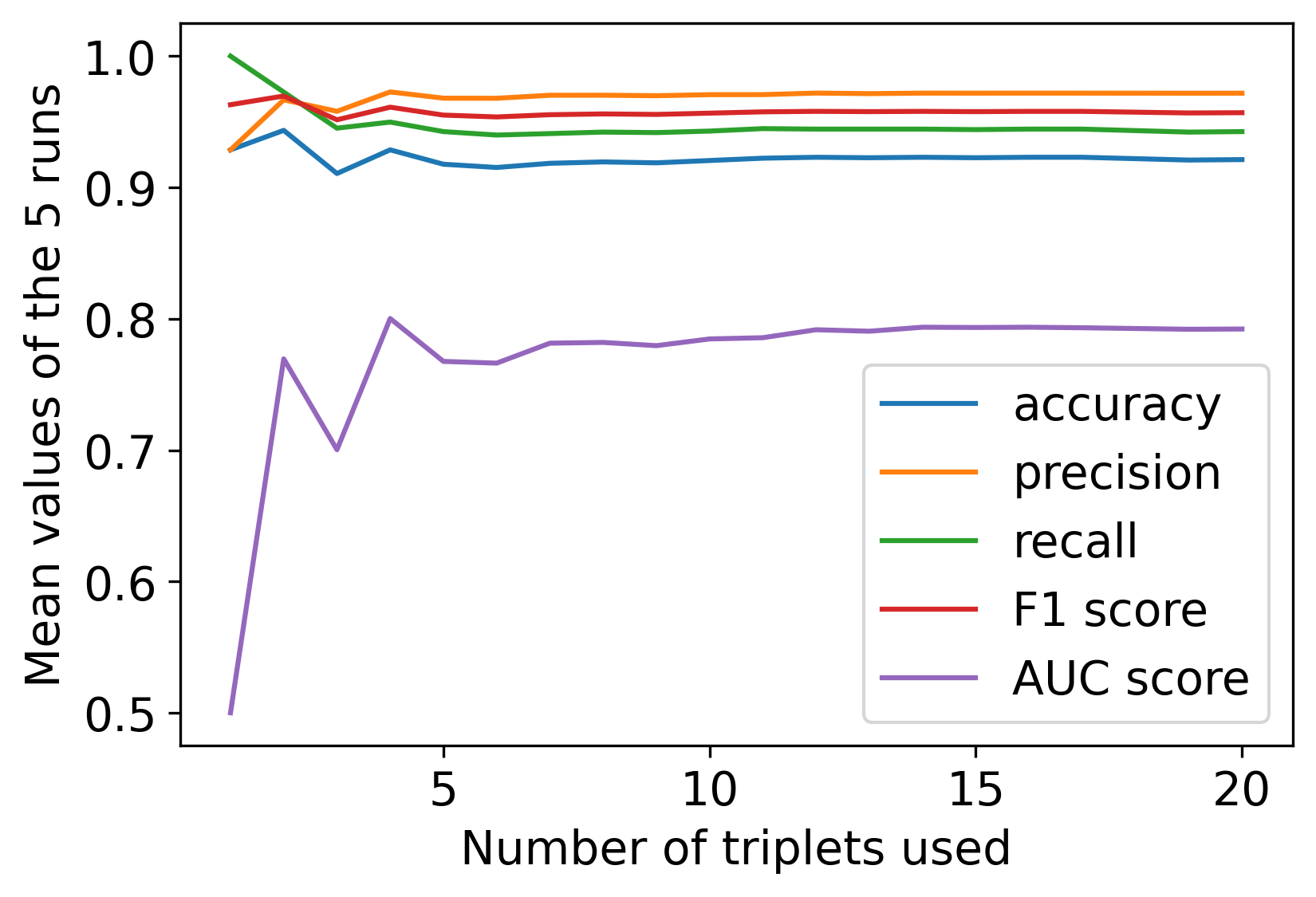}
        \caption{Results of the GNB-A algorithm on the Thyroid dataset}
        \label{subfig:thyroid_gnba}
    \end{subfigure}
    \begin{subfigure}[b]{0.45\textwidth}
        \centering
        \includegraphics[width=0.95\textwidth]{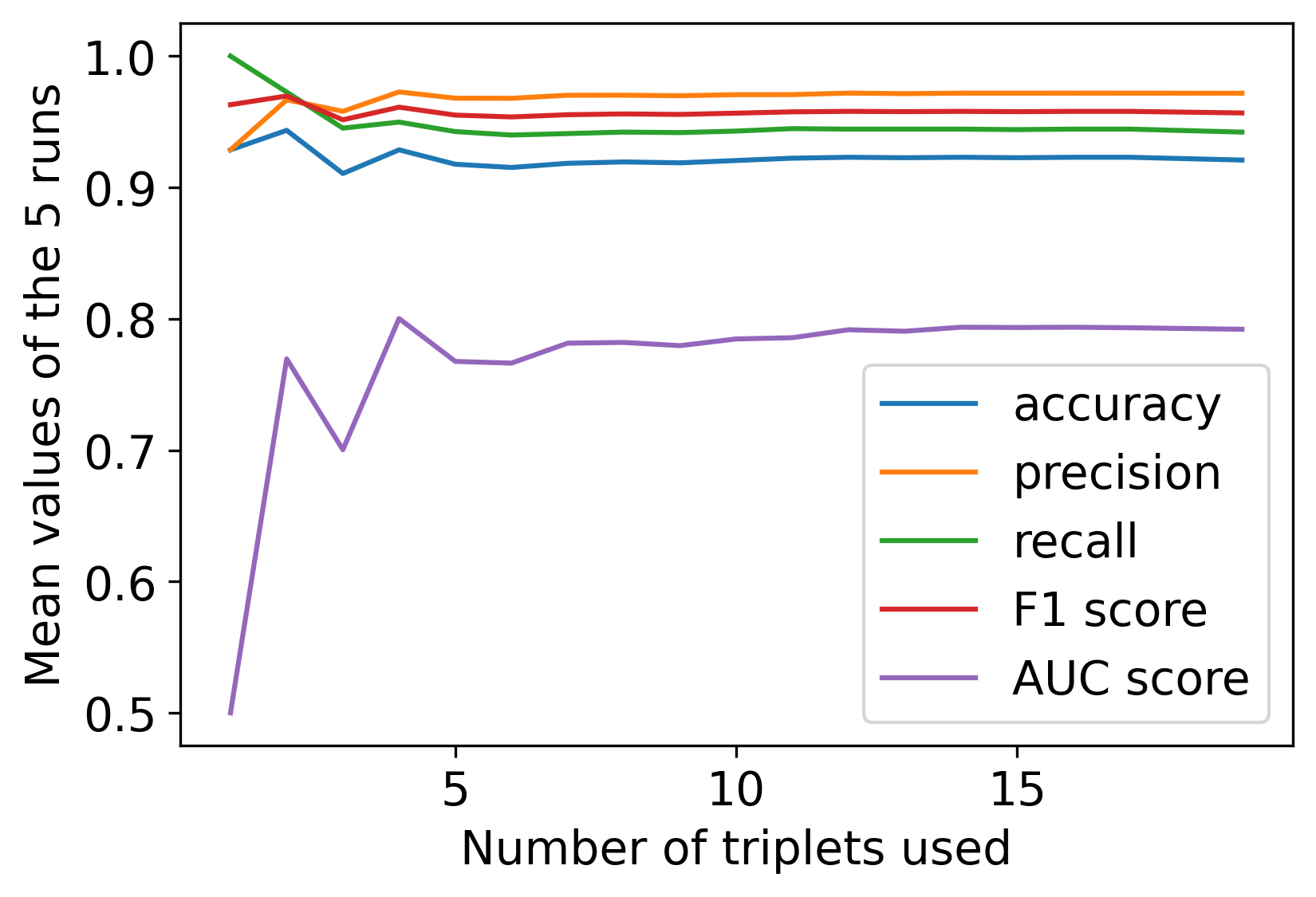}
        \caption{Results of the GNB-O algorithm on the Thyroid dataset}
        \label{subfig:thyroid_gnbo}
    \end{subfigure}
    \caption{}
    \label{fig:resultsThyroidAnnTrain}
\end{figure}

We have discretized the six continuous attributes with the method explained in Subsection~\ref{dataprep}, and joined classes $1$ and $2$ (irregular thryoid level) into a single class. We have then removed column $b14$ as it contained $3771$ $1$'s but only a single $0$. The algorithm was ran $5$ times on it with a random $15\%$ test set selection. The average results of the runs can be seen on Figure \ref{fig:resultsThyroidAnnTrain}.

It seems that for this database specifically, using about $5$ triplets is optimal, as any further triplet introduced does not increase the accuracy, precision, recall or F1 score significantly. It is interesting to note that even by adding new triplets over-fitting is avoided.

\subsubsection{Parkinson dataset}

\begin{figure}[H]
    \centering
    \begin{subfigure}[b]{0.45\textwidth}
        \centering
        \includegraphics[width=0.95\textwidth]{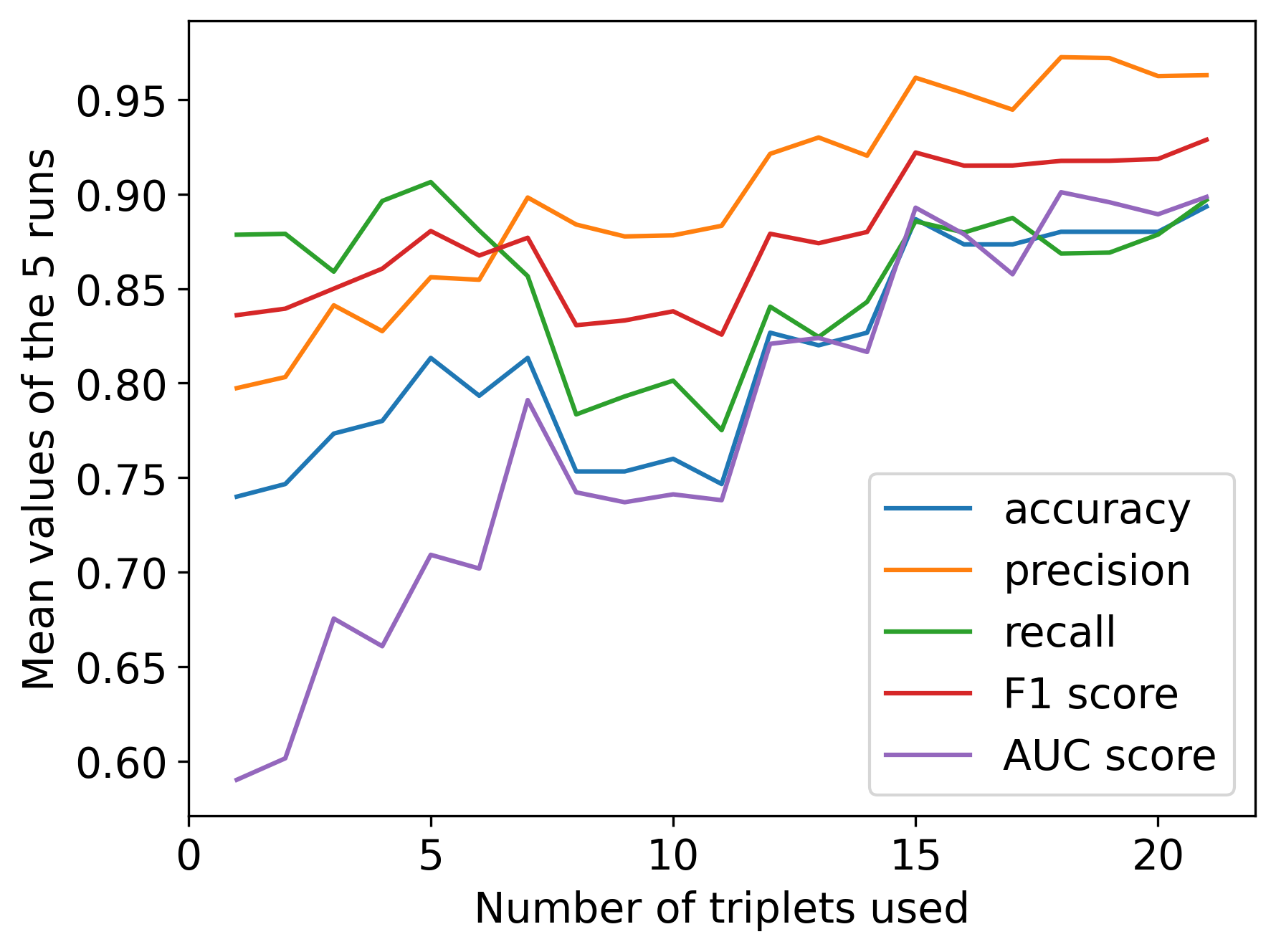}
        \caption{Results of the GNB-A algorithm on the Parkinson dataset}
        \label{subfig:parkinson_gnba}
    \end{subfigure}
    \begin{subfigure}[b]{0.45\textwidth}
        \centering
        \includegraphics[width=0.95\textwidth]{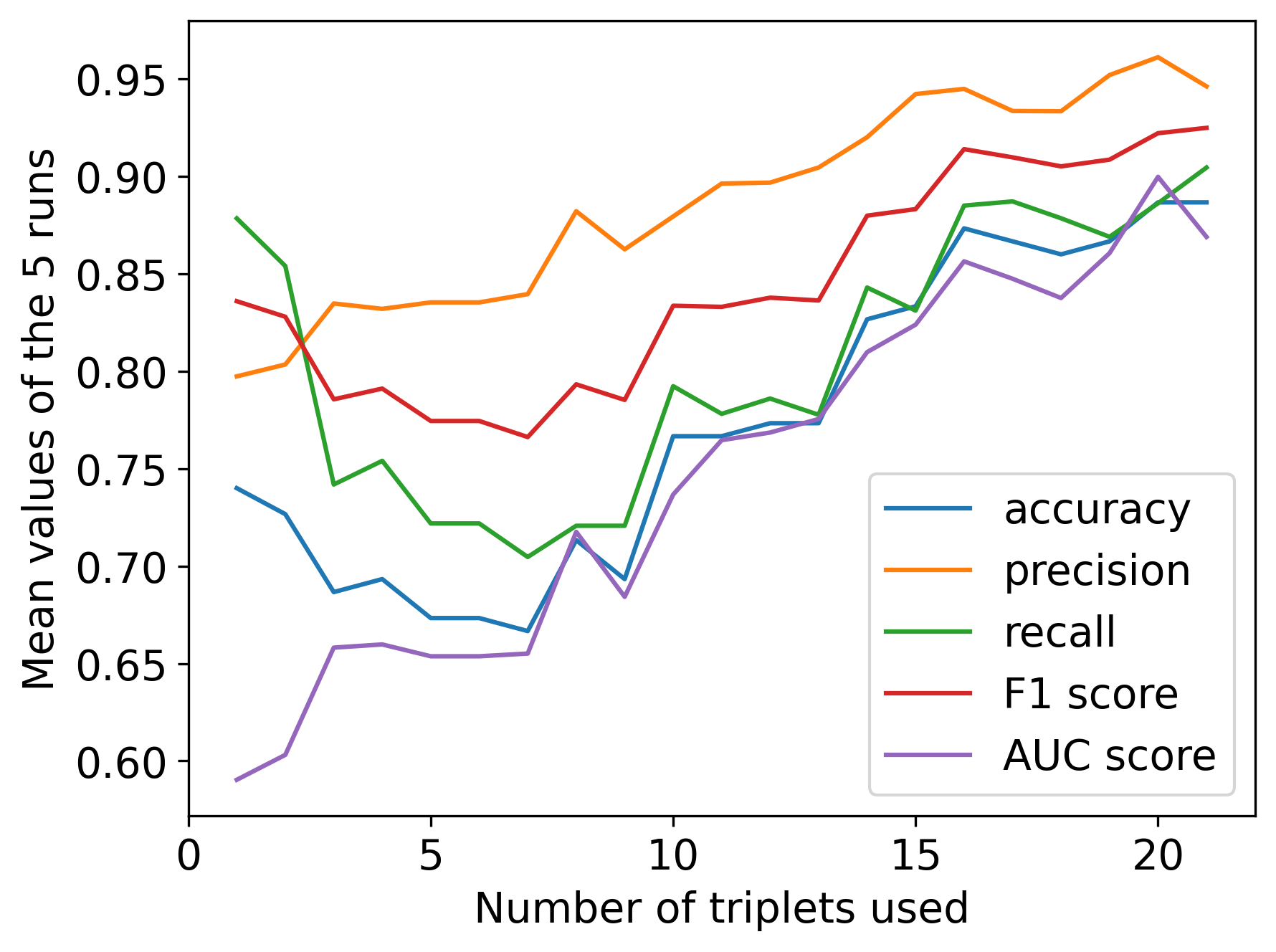}
        \caption{Results of the GNB-O algorithm on the Parkinson dataset}
        \label{subfig:parkinson_gnbo}
    \end{subfigure}
    \caption{}
    \label{fig:resultsParkinson}
\end{figure}

The Parkinsons Disease (PD) Data Set contains biomedical voice measurements from 31 people, 23 with Parkinson's disease. The aim is to discriminate healthy people from those with PD.
The average results of the runs can be seen in Figure \ref{fig:resultsParkinson}.
It is worth observing that the accuracy scores were the highest when almost all features were added.

\subsection{Comparison of GNB-A and GNB-O with other methods}\label{accuracy_comparison}

\begin{table}[H]\label{tab:results_acc}
\centering
\begin{tabular}{|c|c|c|c|c|}
\hline
Dataset\textbackslash{}Model &  GNB-A & GNB-O  &   NB   & TAN \\ \hline
Wdbc                         & 0.9349 & \textbf{0.9442} & 0.8419 & 0.8884 \\
Heart Disease                & 0.7956 & \textbf{0.8133} & 0.7156 & 0.7511 \\
Diabetes                     & 0.8692 & 0.8692 & 0.8692 & \textbf{0.9154} \\
Thyroid Ann Train            & 0.9212 & 0.9208 & 0.9283 & \textbf{0.941} \\ 
Parkinson dataset            & \textbf{0.8933} & 0.8867 & 0.7533 & 0.76 \\ \hline
\end{tabular}
\caption{Accuracy}
\end{table}

\begin{table}[H]\label{tab:results_prec}
\centering
\begin{tabular}{|c|c|c|c|c|}
\hline
Dataset\textbackslash{}Model & GNB-A  &  GNB-O &   NB   & TAN \\ \hline
Wdbc                         & 0.921  & 0.9341 & 0.8688 & \textbf{0.9777} \\
Heart Disease                & 0.8249 & \textbf{0.828} & 0.6913 & 0.7753 \\
Diabetes                     & 0.9419 & 0.9419 & 0.8243 & \textbf{0.9605} \\
Thyroid Ann Train            & \textbf{0.9717} & \textbf{0.9717} &  0.9286 & 0.971 \\  
Parkinson dataset            & \textbf{0.9628} & 0.9462 & 0.8158 & 0.8838 \\ \hline
\end{tabular}
\caption{Precision}
\end{table}

\begin{table}[H]\label{tab:results_rec}
\centering
\begin{tabular}{|c|c|c|c|c|}
\hline
Dataset\textbackslash{}Model &  GNB-A & GNB-O  &   NB   & TAN \\ \hline
Wdbc                         & 0.9238 & \textbf{0.9339} & 0.7316 & 0.7542 \\
Heart Disease                & 0.7262 & \textbf{0.7621} & 0.7214 & 0.6405 \\
Diabetes                     & 0.845 & 0.845 & 0.8261 & \textbf{0.9035} \\
Thyroid Ann Train            & 0.9425 & 0.9421 &  \textbf{0.9996} & 0.9655 \\  
Parkinson dataset            & 0.8968 & \textbf{0.9045} & 0.8718 & 0.7916 \\ \hline
\end{tabular}
\caption{Recall}
\end{table}

\begin{table}[H]\label{tab:results_f1}
\centering
\begin{tabular}{|c|c|c|c|c|}
\hline
Dataset\textbackslash{}Model &  GNB-A & GNB-O  &   NB   &  TAN \\\hline
Wdbc                         & 0.9213 & \textbf{0.9333} & 0.7928 & 0.8501 \\
Heart Disease                & 0.7677 & \textbf{0.7887} & 0.6989 & 0.7008 \\
Diabetes                     & 0.8908 & 0.8908 & 0.8243 & \textbf{0.9305} \\
Thyroid Ann Train            & 0.9569 & 0.9567 &  0.9628 & \textbf{0.9682} \\  
Parkinson dataset            & \textbf{0.9271} & 0.924  & 0.8406 & 0.8315 \\ \hline
\end{tabular}
\caption{F1-score}
\end{table}

\begin{table}[H]\label{tab:results_auc}
\centering
\begin{tabular}{|c|c|c|c|c|}
\hline
Dataset\textbackslash{}Model &  GNB-A & GNB-O  &   NB  &  TAN \\ \hline
Wdbc                         & 0.9346 & \textbf{0.9424} & 0.825 & 0.8713 \\
Heart Disease                & 0.7902 & \textbf{0.8077} & 0.7211 & 0.7392 \\
Diabetes                     & 0.877 & 0.8986 & 0.8598 & \textbf{0.9202} \\
Thyroid Ann Train            & \textbf{0.7922} & 0.7920 & 0.4998 & 0.7852 \\  
Parkinson dataset            & \textbf{0.8984} & 0.8689 & 0.632  & 0.7408 \\ \hline
\end{tabular}
\caption{ROC AUC score}
\end{table}

\section{Conclusion}
This paper introduces the Generalized Naive Bayes structure as a particular case of junction trees, called third-order cherry trees. We gave in the paper two new algorithms for the construction of the GNB structure. GNB-A which is a greedy algorithm, and GNB-O which is the optimal algorithm, under the assumption that the most informative triplet is contained by this structure. 

We proved that the probability distribution associated with the GNB structure gives a better approximation to the real data than the probability distribution associated with the Naive Bayes structure.

We also proved that our GNB-O algorithm gives the optimal GNB- distribution in terms of minimization of the Kullback-Leibler divergence if the triplet with maximum information content is contained by this structure.

Both of our algorithms were tested on real data, and compared with related algorithms, by using different accuracy measures. It turned out that our algorithms work many times better than the related algorithms NB and TAN. 

It is very important to highlight here that our algorithms opposed to the black-box type of algorithms are transparent, overfitting minimization is encoded in the optimization task and it is easy to interpret the results, and the relevant features.

While the optimized algorithm GNB-O is optimal over all of the explanatory variables considered, the greedy algorithm GNB-A finds in each step (adding a new variable) the best choice to minimize the Kullback Leibler divergence. In this way, GNB-A can be stopped at any level, the structure will be a cherry tree with a high weight, whereas GNB-O is optimal when we use the whole structure. 

Based on the GNB structures, feature importance scores were introduced which are independent of the "values" taken on by the attributes, because these are related to the information content concept. Moreover, accuracy scores (accuracy, precision, recall, f1 score, AUC) were assigned to the increasing number of cherries involved. 

We gave a comparison between the algorithms introduced here with the already-existing related algorithms from structural, algorithmic, and complexity points of view, explaining what they have in common and the differences between them. We highlight here that despite the graph of the GNB structure which as graph coincides with the TAN graph structure, the GNB structure is optimized to fit the data by minimizing KL divergence. Moreover, the algorithms introduced, maximize at the same time the added information and minimize the redundancy in each step.

We hope this work will greatly impact to a large range of classification problems related to different fields. In medical studies, it can be a great tool because the results are easy to interpret, and feature importance may have a very positive impact on different studies. The methods introduced here can be easily applied to more general classification tasks (not only binary).
The results presented are based on quantile-discretized to keep it as general as possible. We are sure that all results may be improved by considering other discretizations, but analyzing this was beyond the scope of this paper.

\appendix

%\section{Sample Appendix Section}

 \bibliographystyle{elsarticle-num} 
% \bibliography{cas-refs}





% \end{thebibliography}
\end{document}